%% file: main.tex
\documentclass[letterpaper, 10 pt, conference]{ieeeconf}  

\IEEEoverridecommandlockouts                              

\usepackage{amsmath}

\usepackage{url}

\usepackage{subfigure}
\usepackage{amssymb}
\usepackage{latexsym}
\usepackage{bm}  
\usepackage{mathtools}

\usepackage{todonotes}

\usepackage{mywidebar}

\usepackage{graphicx}
\usepackage{graphics}

\usepackage{float}
\restylefloat{table}

\usepackage{subfigure}
\usepackage{amssymb}
\usepackage{latexsym}
\usepackage{bm}  
\usepackage{mathtools}
\usepackage{adjustbox}
\usepackage{todonotes}

\usepackage{mywidebar}

\usepackage{booktabs}

\usepackage{url}
\usepackage{breakurl}
\makeatletter
\let\NAT@parse\undefined
\makeatother
\usepackage[colorlinks,citecolor=blue,urlcolor=blue,breaklinks]{hyperref}

\newcommand*{\Scale}[2][4]{\scalebox{#1}{$#2$}}%
\newcommand{\scaleG}[1]{\Scale[0.7]{#1}}

\title{\LARGE \bf
PLVS: A SLAM System with Points, Lines, Volumetric Mapping, and 3D Incremental Segmentation
}

\author{Luigi Freda$^{1}$ 
\thanks{$^{1}$
        {\tt\small luigifreda$[@]$gmail.com}}%
        }

\begin{document}

\maketitle
\thispagestyle{empty}
\pagestyle{empty}

\begin{abstract}
\input{sections/abstract}
\end{abstract}

\section{Introduction}
\input{sections/introduction}

\section{Overview}
\label{Sect:Overview}

\input{sections/system-overview}

\section{Sparse SLAM}
\label{Sect:SparseSLAM}
\input{sections/sparse-slam}

\section{Volumetric Mapping}
\label{Sect:VOMA}

\input{sections/voma}

\section{Incremental Segmentation}\label{Sect:Segmentation}

\input{sections/segmentation}


\section{Experimental Results}
\label{Sect:ExperimentalResults}
\input{sections/exp-results-summary}

\section{Conclusions}
\label{Sect:Conclusions}

\input{sections/conclusions}

\appendices

\section{}
\input{sections/appendix}







\bibliographystyle{IEEEtran}
\bibliography{main}

\end{document}

%% file: sections/abstract.tex
This document presents PLVS: a real-time system that leverages sparse SLAM, volumetric mapping, and 3D unsupervised incremental segmentation. PLVS stands for Points, Lines, Volumetric mapping, and Segmentation. It supports RGB-D and Stereo cameras, which may be optionally equipped with IMUs. The SLAM module is keyframe-based, and extracts and tracks sparse points and line segments as features. Volumetric mapping runs in parallel with respect to the SLAM front-end and generates a 3D reconstruction of the explored environment by fusing point clouds backprojected from keyframes. Different volumetric mapping methods are supported and integrated in PLVS. We use a novel reprojection error to bundle-adjust line segments. This error exploits available depth information to stabilize the position estimates of line segment endpoints. An incremental and geometric-based segmentation method is implemented and integrated for RGB-D cameras in the PLVS framework. We present qualitative and quantitative evaluations of the PLVS framework on some publicly available datasets. The appendix details the adopted stereo line triangulation method and provides a derivation of the Jacobians we used for line error terms. The software is available as open-source. 

%% file: sections/introduction.tex
\begin{figure*}[h!]
\centering
\subfigure[]{\includegraphics[scale=0.225]{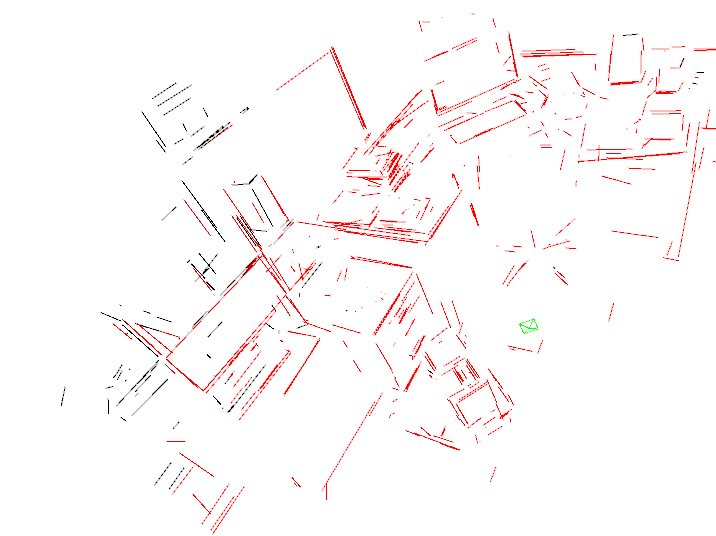}}
\subfigure[]{\includegraphics[scale=0.225]{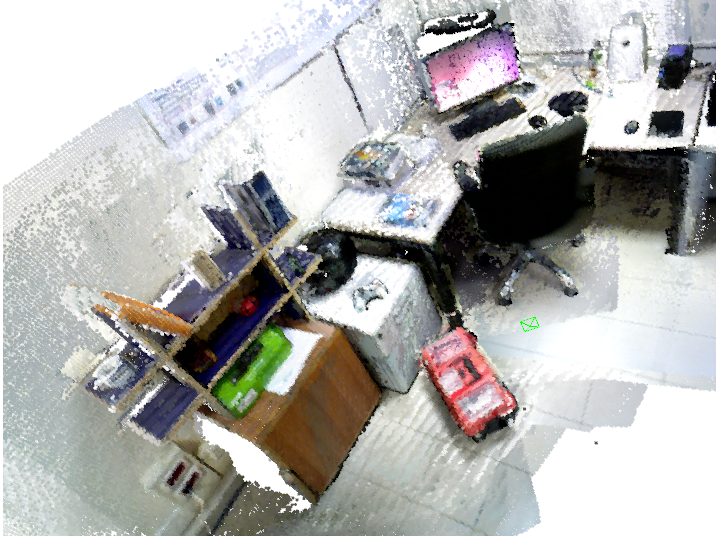}}
\subfigure[]{\includegraphics[scale=0.225]{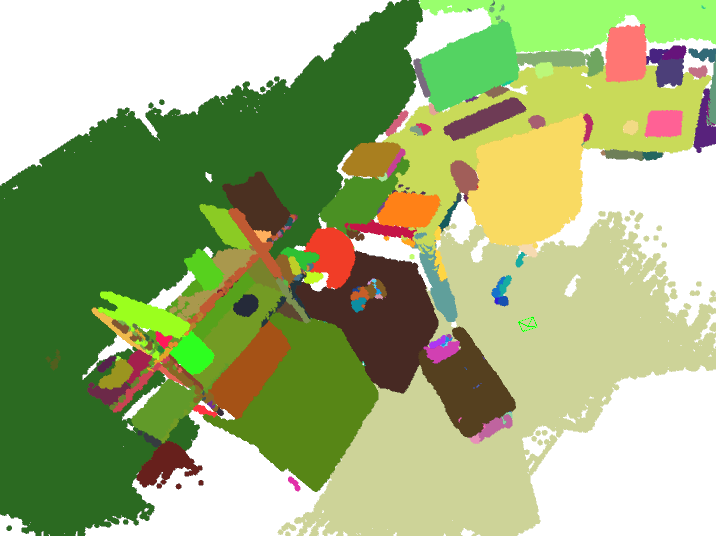}}
\caption{Three aspects of a 3D reconstruction (a) line segments (b) point cloud (c) segmented point cloud. For simplicity, we do not show point features. The volumetric map has been obtained by using an octree-based model with 1cm resolution.}
\label{Fig:PLVSMain}
\end{figure*}

\begin{figure*}[!th]
\centering
\subfigure[]{\includegraphics[scale=0.26]{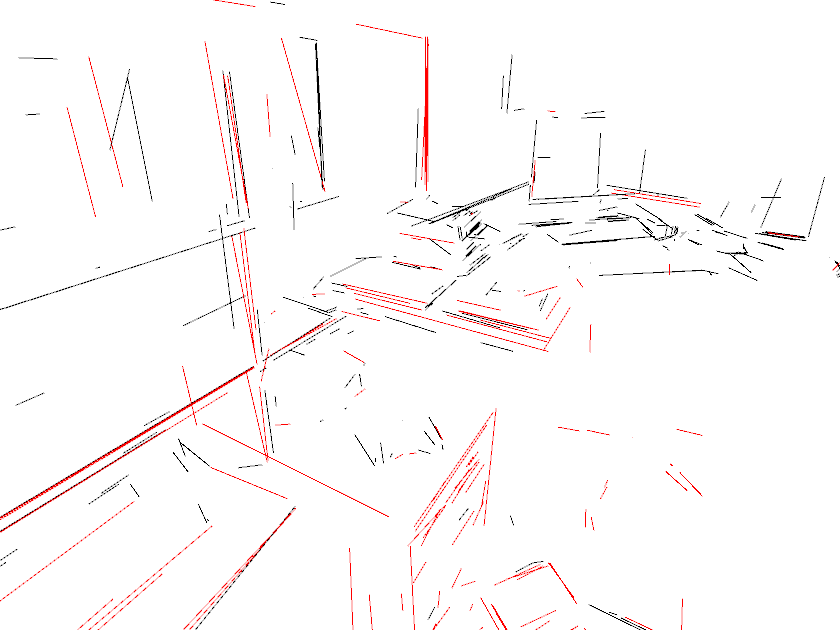}}\quad
\subfigure[]{\includegraphics[scale=0.26]{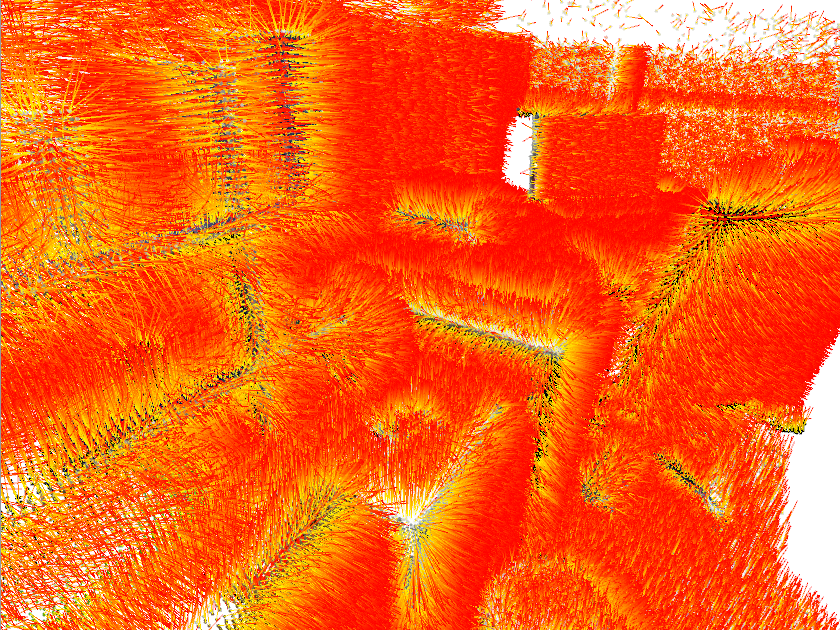}}\\
\subfigure[]{\includegraphics[scale=0.26]{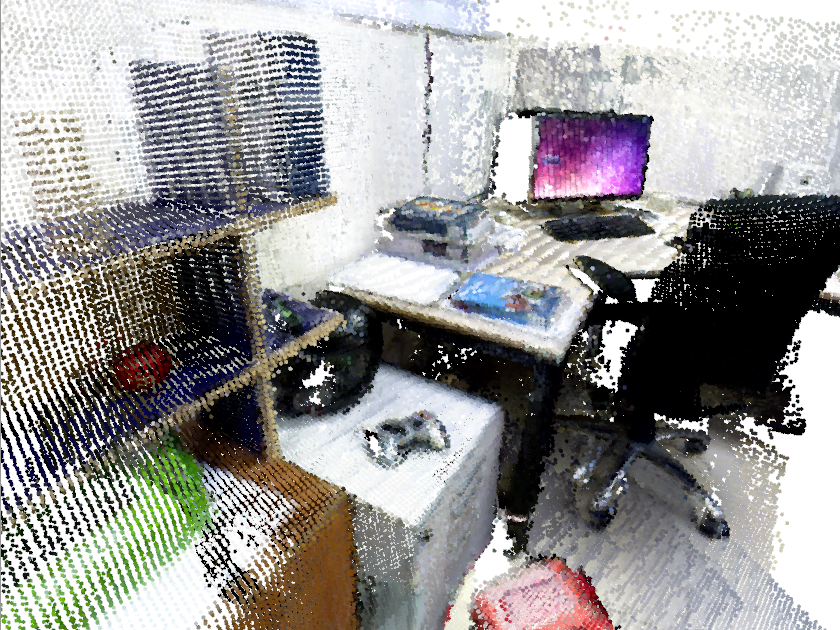}}\quad
\subfigure[]{\includegraphics[scale=0.26]{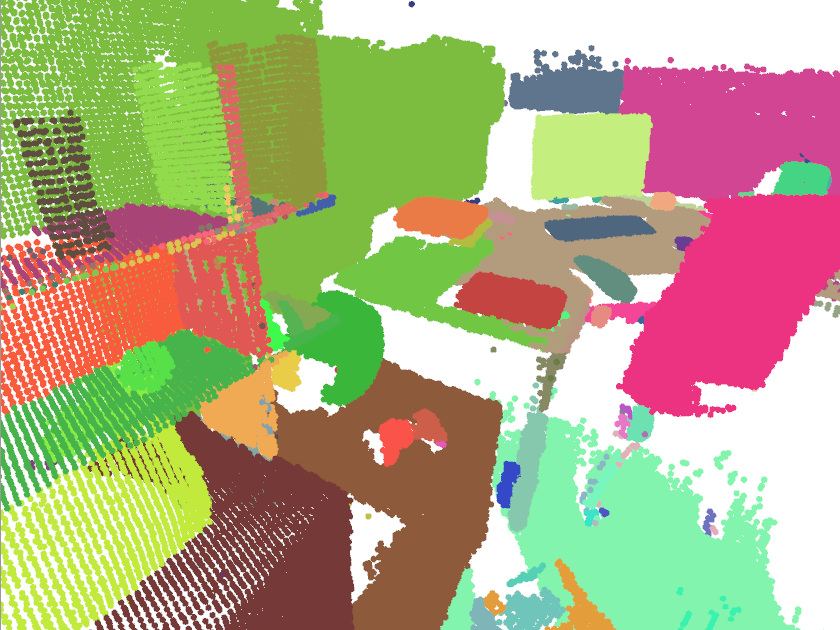}}
\caption{Details of the 3D reconstruction of Fig.~\ref{Fig:PLVSMain} (a) line segments (b) normals (c) point cloud (d) segmented point cloud.}
\label{Fig:PLVSDetail}
\end{figure*}

The goal of the PLVS project is to create a robust and agile SLAM framework that can reliably run onboard of small-sized robotic platforms and generate consistent dense maps of the surrounding environment. The target sensors are RGB-D and stereo cameras, that may be optionally equipped with IMU devices. 
   
The following is a list of the main PLVS features. 
\begin{itemize}
\item PLVS is an open-source framework which (ii) exploits both keypoints and keylines for camera tracking and sparse mapping (ii) integrates different methods for volumetric mapping: in particular, octrees, octomap, and spatially-hashed voxels with Truncated Signed Distance Fields (TSDFs) and meshes. 
\item It is designed to work with RGB-D and stereo cameras, with and without IMU devices.
\item A novel reprojection error for line segments is implemented in the framework. This error exploits available depth information to stabilize the position estimates of line segment endpoints. Additionally, it is general and can be used with both RGB-D and stereo camera systems. 
\item It comes with an incremental method for segmenting the volumetric map. This method is geometry-based and unsupervised. At present time, it only works with RGB-D cameras and takes advantage of the extracted line segment information generated by SLAM.
\item The system runs entirely on the CPU. Optionally, keypoint extraction can be moved to GPU to reduce the computational load.  
\end{itemize}

PLVS is a modular and versatile system. Its different capabilities are organized into divisions that can be enabled/disabled and configured by users in different ways. This allows to finely trade-off map accuracy/resolution versus CPU load and to adapt the framework capabilities to the system at hand. Figures~\ref{Fig:PLVSMain}-\ref{Fig:PLVSDetail} show the details of a map obtained live with PLVS. 
An open-source implementation is available at \href{https://github.com/luigifreda/plvs}{github.com/luigifreda/plvs}.

This document is organized as follows. First, an overview of the system is provided in Sect.~\ref{Sect:Overview}. Next, Sect.~\ref{Sect:SparseSLAM} describes the sparse SLAM subsystem. Then, we provide an overview of the volumetric mapping (Sect.~\ref{Sect:VOMA} and incremental segmentation modules (Sect.~\ref{Sect:Segmentation}). Finally, we report some qualitative and quantitative results in Sect.~\ref{Sect:ExperimentalResults}.

%% file: sections/system-overview.tex
At high-level, the PLVS system can be divided in two main components: \emph{(i)} the sparse SLAM and \emph{(ii)} the VOlumetric MApping (VOMA).

We built the sparse SLAM subsystem upon ORB-SLAM~\cite{mur2017orb, campos2021orb}. It is keyframe-based. It detects, tracks and maps points and line segments as sparse features. PLVS uses sparse SLAM  to accurately localize camera and select relevant keyframes. Indeed, PLVS is available in two different versions: the first one, PLVS I, is based on ORB-SLAM2, while the newest one, PLVS II, is based on ORB-SLAM3 and support camera systems equipped with IMUs.

The VOMA runs in a parallel thread and uses the SLAM keyframes, with their estimated camera positions and backprojected point clouds, to build a 3D reconstruction of the explored environment. Different methods are supported and integrated in the system in order to fuse keyframe point clouds (see Sect.~\ref{Sect:VOMA}). An incremental 3D segmentation method is implemented in the VOMA for RGB-D cameras (see Sect.~\ref{Sect:Segmentation}).

Indeed, two kinds of maps are built in parallel:
\begin{itemize}
\item the \emph{sparse map} ${\cal M}_s = ({\cal K},{\cal P},{\cal L})$, which is the output of the sparse SLAM, includes a set $\cal K$ of keyframes and the sets ${\cal P}$ and $\cal L$, which respectively represent the 3D points and the 3D line segments corresponding to features extracted and matched in the keyframes in $\cal K$;
\item the \emph{volumetric map} ${\cal M}_v$, which is built by the VOMA, results from the "integration" of the point clouds backprojected from the keyframes $\cal K$ (see Sect.~\ref{Sect:VOMA}). 
\end{itemize}
In order to maintain consistency between the two maps, the volumetric map is re-built (or adjusted) whenever the SLAM map or its underlying pose graph go through a global adjustment (e.g. when loop closures are detected).  

The PLVS framework runs entirely on CPU with the aim of being portable on small-size mobile robots (which do not necessarily offer advanced GPU capabilities). It is modular, which means its different divisions can be enabled /disabled and configured in many ways by users. This allows to finely trade-off map accuracy/resolution versus CPU load and adapt the framework to the system at hand.

%% file: sections/sparse-slam.tex

In PLVS, the architecture of the sparse SLAM system  reflects the one of ORB-SLAM~\cite{mur2017orb, campos2021orb}. The extraction of keylines on images, and the tracking, mapping, bundle-adjustment of 3D line segments proceed side by side with the corresponding operations on keypoints and 3D points.

Below, we provide futher details about the management, reprojection errors, and adopted representations for keylines and 3D line segments.

\subsection{Notation}\label{Sect:Notation}

Throughout this paper, we use the following notation. $\Omega \subset \mathbb{R}^2$ is the image domain. The generic $i$-th RGB-D image $(I_i,D_i)$ consists of a rectified colour image $I_i: \Omega \rightarrow \mathbb{N}^3$ and a registered depth image $D_i: \Omega \rightarrow \mathbb{R}^+$. The depth image $D_i$ returns for each pixel $\bm{u}=[u,v]^T \in  \Omega$ the positive  distance between the image plane and the closest object point intercepted by the ray emanated from the camera center and passing through $\bm{u}$. 

The camera transformation matrix corresponding to $(I_i,D_i)$ is denoted by $\textbf{T}_i = \in SE(3)$, transforming a point
from the world frame to the camera frame. One has 
\begin{equation}
\textbf{T}_i = \begin{bmatrix}
       \textbf{R}_i &   \bm{t}_i       \\
       \textbf{0}^T & 1 
     \end{bmatrix}
\end{equation}
where $\textbf{R}_i \in SO(3)$ and $\bm{t}_i \in \mathbb{R}^3$. For 3D points, we use the superscripts $c$ and $w$ to indicate the camera and world frames in which point coordinates are expressed respectively. 

Given a vector $\bm{v}$, we define the vector $\widebar{\bm{v}} \triangleq [\bm{v}^T,1]^T$. 
Moreover, we define  the operator $\bm{\gamma}: \mathbb{R}^3 \rightarrow \Omega$ as $\bm{\gamma}([u,v,w]^T)=[u/w,v/w]^T$. 

Let $\textbf{K}\in \mathbb{R}^{3\times 3}$ be the fixed camera calibration matrix. The camera projection $\bm{u} \in  \Omega$ of a 3D point $\bm{P}^c = [x,y,z]^T \in \mathbb{R}^3$ is
\begin{equation}\label{eq:CameraProjection}
\bm{u} = \bm{\gamma}(\textbf{K} \bm{P}^c) = \Bigg[ f_x \frac{x}{z} + c_x, f_y \frac{y}{z} + c_y \Bigg]^T
\end{equation}
where $(f_x,f_y) \in \mathbb{R}^2$ defines the focal length and $(c_x,c_y) \in \mathbb{R}^2$ is the principal point. 

The 3D backprojection of a pixel $\bm{u} \in \Omega$ given the depth image $D_i$ is 
\begin{equation}\label{Eq:Backprojection}
\bm{P} = \textbf{K}^{-1}  \widebar{\bm{u}} D_i(\bm{u}) = \Bigg[ \frac{(u-c_x)}{f_x}\delta, \frac{(v-c_y)}{f_y}\delta, \delta \Bigg]^T
\end{equation}
where $\delta = D_i(\bm{u}) \in \mathbb{R}^+$. 

For an RBG-D image $(I_i,D_i)$, we define (with a slight abuse of notation) the projection map $\bm{\pi}_i : \mathbb{R}^3 \rightarrow \Omega $ and 
the backprojection map $\bm{\beta}_i :  \Omega \rightarrow \mathbb{R}^3$  as follows
\begin{equation}
\bm{\pi}_i(\bm{P}^w)\triangleq \bm{\gamma}(\textbf{K} [\textbf{R}_i,\bm{t}_i]\widebar{\bm{P}}^w),
\qquad \bm{\beta}_i(\bm{u}) = \textbf{K}^{-1}  \widebar{\bm{u}} D_i(\bm{u}).
\end{equation}

On the image plane, we consider 2D line equations of the following form 
\begin{equation}\label{Eq:LineEquation}
\bm{n}^T \bm{u} - h = 0 
\end{equation} 
where $\bm{n} = [n_u,n_v] \in \mathbb{R}^2$ is the unit normal direction and $h \in \mathbb{R}$ is the signed distance of the considered line from the origin. Given two points $\bm{u}_1, \bm{u}_2 \in \Omega$, we compute $\bm{n}=[\Delta v, -\Delta u]^T/\sqrt(\Delta^2 u + \Delta^2 v)$ and $h=\bm{n}^T\bm{u}_1$ (or $h=\bm{n}^T\bm{u}_2$). Alternatively, we can compute the homogeneous representation of the 2D line $[l_1,l_2,l_3]^T = \widebar{\bm{u}}_1 \times \widebar{\bm{u}}_2$ and an equivalent line equation is obtained as: $[l_1,l_2,l_3]^T \widebar{\bm{u}} = 0$~\cite{hartley2003multiple}.

Given a covariance matrix $\bm{\Sigma}  \in \mathbb{R}^{n \times n}$, the \textit{Mahalanobis distance} between two vectors $\bm{v},\bm{w} \in \mathbb{R}^n$ is $\| \bm{v} - \bm{w}\|_{\bm{\Sigma}} = \big((\bm{v} - \bm{w})^T\bm{\Sigma}^{-1} (\bm{v} - \bm{w})\big)^{1/2}$. We denote by $\textbf{I}_n$ the identity matrix of size $n \times n$.

\input{sections/points}

\begin{figure}[!t]
\begin{center}
{\centering\includegraphics[scale=1]{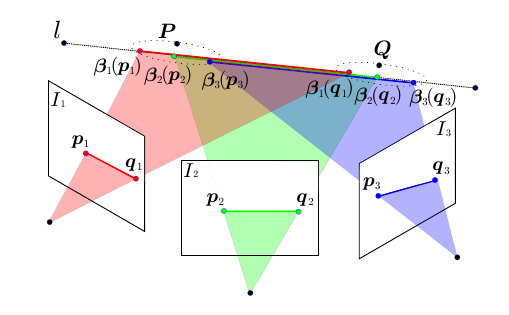}} \caption{The schema shows how three line segment observations $\{(\bm{p}_i,\bm{q}_i)\}_{i=1}^{3}$ of the same 3D line segment $l$ are  backprojected onto the 3D line at different elements of the class ${\cal C}_l$.
} \label{Fig:LineRepresentation}
\end{center}
\end{figure}

\subsection{Line Segments Representation}

A 3D line segment $l$ is represented in the sparse map ${\cal M}_s$ by a pair of 3D endpoints $\bm{P}, \bm{Q} \in \mathbb{R}^3$. 
The detection of $l$ in an image (following a 2D line fitting stage) consists of a support region~\cite{ma2012invitation}, which is a 2D compact and connected region ${\cal W}(l) \subset \Omega$ represented  by two endpoints $\bm{p}_i,\bm{q}_i \in \Omega$. The matched pair of image endpoints $(\bm{p}_i,\bm{q}_i)$ is also referred to as the $i$-th \emph{line segment observation} of $l$.  

Due to noise, possible occlusions, and other factors, the 3D line segment $l$ may not be fully observed by the camera and endpoints are often inconsistent across images~\cite{liu20233d}. Furthermore, a 3D line segment can be fragmented into multiple line segments (belonging to the same line in 3D). In fact, backprojecting all observations $\{(\bm{p}_i,\bm{q}_i)\}_{i=1}^{n_l}$ of $l$ one obtains a class of 3D line segments ${\cal C}_l =  \{(\bm{\beta}_i(\bm{p}_i),\bm{\beta}_i(\bm{q}_i)) \in \mathbb{R}^3 \times \mathbb{R}^3, i=1,...,n_l \}$ where, in general, $\| \bm{\beta}_i(\bm{p}_i) - \bm{\beta}_i(\bm{q}_i) \| \neq \| \bm{\beta}_j(\bm{p}_j) - \bm{\beta}_j(\bm{q}_j) \|$ for $i \neq j$ (see Fig.~\ref{Fig:LineRepresentation}). Here, for simplicity, we assumed the $i$-th observation $(\bm{p}_i,\bm{q}_i)$ occurs in the $i$-th image $(I_i,D_i)$.

We represent a 3D line segment $l$ by using the following elements:
\begin{enumerate}
\item the 2D image endpoints $\bm{p}_i,\bm{q}_i \in \Omega$ in the generic image $(I_k,D_k)$ where $l$ is detected and successfully matched;
\item the 3D backprojected endpoints $\bm{\beta}_k(\bm{p}_i),\bm{\beta}_k(\bm{q}_i) \in \mathbb{R}^3$ \emph{attached} to the camera frame of $(I_k,D_k)$;
\item the 3D map endpoints $\bm{P}, \bm{Q} \in \mathbb{R}^3$ which represent the line segment $l$ in the map ${\cal M}_s$. 
\end{enumerate}
For simplicity, we do not explicit the implied mapping $k=k(i)$ which is built at extraction/matching time and associates the $i$-th observation of $l$ with the $k$-th frame/keyframe. 

A line segment observation $(\bm{p}_i,\bm{q}_i)$ is defined \emph{stereo} if both the depths $D_k(\bm{p}_i)$ and $D_k(\bm{q}_i)$ exist and are finite\footnote{Indeed, with commodity RGB-D sensors, the depth $D_i$ may be not defined for each pixel.}; \emph{mono} otherwise. Only a stereo observation allows to compute $\bm{\beta}_k(\bm{p}_i)$ and $\bm{\beta}_k(\bm{q}_i)$. The availability of the 3D backprojected points allows the introduction of a new type of error which takes into account depth information to better estimate the positions of the line segment endpoints during bundle adjustment stages (see Sect.~\ref{Sect:ReprojectionErrors}). 

In general, a RGB-D camera provides the depths $D_k(\bm{p}_i)$ and $D_k(\bm{q}_i)$. In the case a binocular camera or when triangulating line segments belonging to different camera images, depths can be estimated in closed form by using the method described in the appendix (Sect.~\ref{Sect:LineTriangulation}).

\subsection{Line Segments Detection and Matching}

For each processed image $I_i$, a Gaussian image pyramid is first computed. Then, line segments are extracted on each level of the image pyramid by using the EDlines method described in~\cite{akinlar2011edlines}. This consists of a linear time algorithm which also includes a false detection control. We selected the EDlines method for its repeatability, precision, and efficiency. 

On each level of the image pyramid, we model the uncertainty in the endpoints detection process with a specific noise variance. We denote by $\sigma^2_{li} \in \mathbb{R}$ the noise variance on each coordinate of the image endpoints $\bm{p}_i,\bm{q}_i$.
 
In order to attain frame-to-frame line matching, a binary descriptor is associated to each extracted line segment by using the Line Band Descriptor (LBD) method~\cite{zhang2013efficient}. 

We use a tiling technique in order to speed up line segment matching. In particular, at extraction time, we compute for each line segment its normal orientation $\theta = {\rm atan2}(n_v,n_u)$ and signed distance $h$. We partition the 2D parameter manifold $(\theta,h)$ into tiles. As a consequence, for each image $I_i$, we group the extracted line segments into their corresponding tiles and store the resulting partition in~$I_i$.

At matching time, we first project a 3D line segment $l \in {\cal M}_s$ over the target image $I_i$; then, we compute its projected line representation $(\theta_l,h_l)$; next, we identify as candidate matches of $l$ in $I_i$ all the line segments of $I_i$ whose parameters $(\theta,h)$ lie in the same tile. Finally, the identified candidate matches are scored according to the Hamming distances between their descriptors. To reject possible false matches, we validate the found nearest neighbor if \emph{(i)} its Hamming distance ratio to the second-closest neighbor is below a certain threshold (in general $0.8$) \emph{(ii)} its line to line distance on the image plane is below a certain threshold.

\begin{figure}[!t]
\begin{center}
{\centering\includegraphics[scale=1.3]{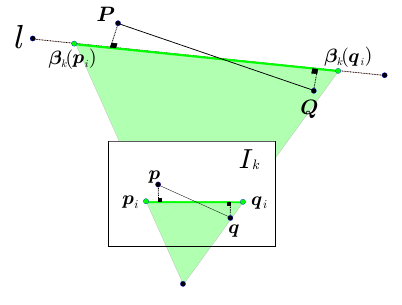}} \caption{The generic 3D line segment $l$, its obversation $(\bm{p}_i,\bm{q}_i)$ in the image $(I_k,D_k)$ and the map endpoints $(\bm{P},\bm{Q})$ with their camera projections $(\bm{p},\bm{q}) = (\bm{\pi}_k(\bm{P}),\bm{\pi}_k(\bm{Q}))$. The 2D reprojection error $e_{2D}$ in eq.~(\ref{Eq:2DReprojectionError}) combines the depicted perpendicular point-line distances of $\bm{p}$ and $\bm{q}$ from the line $(\bm{p}_i,\bm{q}_i)$. On the other hand, the 3D backprojection error $e_{3D}$ in eq.~(\ref{Eq:3DBackprojectionError}) combines \emph{(i)} the 3D perpendicular point-line distances of $\bm{P}$ and $\bm{Q}$ from the 3D line $(\bm{\beta}_k(\bm{p}_i),\bm{\beta}_k(\bm{q}_i))$  and \emph{(ii)} the 3D endpoint-endpoint distances $\|\bm{P}-\bm{\beta}_k(\bm{p}_i)\|$ and $\|\bm{Q}-\bm{\beta}_k(\bm{q}_i)\|$.} \label{Fig:LineErrors}
\end{center}
\end{figure}

\subsection{Line Reprojection and Backprojection Errors}\label{Sect:ReprojectionErrors}

To incorporate line segments into  Bundle Adjustment (BA) we need to define a proper line reprojection error and to compute its differential with respect to both camera pose parameters and 3D line segment parameters. 

Consider the generic line segment $l$ with its $n_l \in \mathbb{N}$ observations $\{(\bm{p}_i,\bm{q}_i)\}_{i=1}^{n_l}$ and its 3D representation $(\bm{P}, \bm{Q})$ in the sparse map ${\cal M}_s$. In BA, each observation $(\bm{p}_i,\bm{q}_i)$ matches $l$ in a precise keyframe $k$ whose camera transformation is $\textbf{T}_k$. 

Given the observation $(\bm{p}_i,\bm{q}_i)$, its equation parameters $(\bm{n}_i,h_i)$ follow from eq.~(\ref{Eq:LineEquation}). On the image plane, we compute the 2D distance  between the camera projection of a 3D point $\bm{X}^w \in \mathbb{R}^3$  and the 2D line passing through the points $\bm{p}_i$ and $\bm{q}_i$ as:
\begin{equation}\label{Eq:2DDistancePointLineSegment}
d_{2D}(i,k,\bm{X}^w) \triangleq \bm{n}_i^T \bm{\pi}_k(\bm{X}^w) - h_i.
\end{equation}
We define the 2D line-line distance vector as:
\begin{equation}\label{Eq:2DDistanceMapPointsLineSegment}
\bm{d}_{2D}(i,k) \triangleq \begin{bmatrix}
      d_{2D}(i,k,\bm{P}^w) \\
      d_{2D}(i,k,\bm{Q}^w)
     \end{bmatrix} \in \mathbb{R}^2.
\end{equation}
Its $2 \times 2$  covariance matrix covariance  matrix is:
\begin{equation}\label{Eq:SigmaD2DLine}
\bm{\Sigma}_{li} =  
\begin{bmatrix}    
\sigma_{d2D}^2(i, k,\bm{P}^w) & 0 \\
 0  & \sigma_{d2D}^2(i, k,\bm{Q}^w)
 \end{bmatrix} \in \mathbb{R}^{2 \times 2}
\end{equation}
\begin{equation}
\sigma_{d2D}^2(i, k,\bm{X}^w) = \Scale[1]{ 
\frac{\partial d_{2D}}{\partial (\bm{p}_i,\bm{q}_i)}\sigma_{li}^2 \bm{I}_4           
\frac{\partial d_{2D}}{\partial (\bm{p}_i,\bm{q}_i)}^T
}
\in \mathbb{R}
\end{equation}
where $\frac{\partial d_{2D}}{\partial (\bm{p}_i,\bm{q}_i)} \in \mathbb{R}^{1 \times 4}$ is the Jacobian of eq.~(\ref{Eq:2DDistancePointLineSegment}) with respect to both image endpoints coordinates. Here, for simplicity, we imposed $\bm{\Sigma}_{li}$ is diagonal.

Given a \textit{mono} line segment observation, its 2D reprojection error is defined as:
\begin{align}\label{Eq:2DReprojectionError}
e_{2D}^L(i, k) &\triangleq   \rho_{\tau_2} \bigg( \| \bm{d}_{2D}(i,k) \|_{\bm{\Sigma}_{li}}\bigg)
\end{align}
where $\rho$ is the robust Huber cost function~\cite{hartley2003multiple, huber2011robust}. Alternatively, in our implementation, we allow users to optionally select a Cauchy cost function.  

It is worth noting that eq.~(\ref{Eq:2DReprojectionError}) combines the 2D point-line distances on the image plane as shown in Fig.~\ref{Fig:LineErrors}.
In our experiments, we have observed better results when BA only includes mono line segments with at least $n_l \geq 3$ observations. 

Now, moving to 3D, 
we define the following 3D distances
\begin{align}\label{Eq:3DDistancePointLineSegment}
d_{3D}(i, k, \bm{X}^w) &\triangleq \frac{\Vert (\bm{X}^c - \bm{\beta}_k(\bm{p}_i))\times (\bm{X}^c - \bm{\beta}_k(\bm{q}_i)) \Vert }{\Vert \bm{\beta}_k(\bm{p}_i) - \bm{\beta}_k(\bm{q}_i) \Vert } \\
\label{Eq:3DDistancePointBackProjPoint}
d_P(\bm{x}, k, \bm{X}^w) &\triangleq \Vert \bm{X}^c - \bm{\beta}_k(\bm{x}) \Vert \\
\label{Eq:CoordinatesFromWtoC}
\bm{X}^c &= \textbf{R}_k \bm{X}^w +\bm{t}_k
\end{align}
where  eq.~(\ref{Eq:3DDistancePointLineSegment}) represents the 3D perpendicular distance between the map point $\bm{X}$ and the 3D line $(\bm{\beta}_k(\bm{p}_i),\bm{\beta}_k(\bm{q}_i))$ (here, $\bm{X}^w$ is transformed to camera frame by using eq.~(\ref{Eq:CoordinatesFromWtoC})) and eq.~(\ref{Eq:3DDistancePointBackProjPoint}) represents the distance between $\bm{X}$ and the backprojection of its associated image point $\bm{x} \in \Omega$. 


Proceeding as above, we define the 3D \textit{backprojection distance} vector:
\begin{equation}\label{Eq:3DDistanceBB}
\bm{d}_B(i, k) \triangleq \begin{bmatrix}
      d_{3D}(i, k,\bm{P}^w) + \mu d_P(\bm{p}_i, k,\bm{P}^w)\\
      d_{3D}(i, k,\bm{Q}^w) + \mu d_P(\bm{q}_i, k,\bm{Q}^w)
     \end{bmatrix} \in \mathbb{R}^2
\end{equation}
where the scalar weight $\mu \in [0,1]$ is used for assigning a different importance to $d_P$. In general, we privilege minimizing $d_{3D}$ distances (i.e. $\mu < 1$).
The $2\times2$  covariance matrix of $\bm{d}_B(i, k)$ is set as
\begin{equation}\label{Eq:SigmaLi}
\bm{\Sigma}_{Li} =  
\begin{bmatrix}
\sigma_{dB}^2(i, k,\bm{P}^w)          
& \bm{0} \\
\bm{0} & 
\sigma_{dB}^2(i, k,\bm{Q}^w)           
\end{bmatrix}  \in \mathbb{R}^{2 \times 2}
\end{equation}  
\begin{equation}\label{Eq:SigmaD3D}
\sigma_{dB}^2(i, k,\bm{X}^w) 
 \triangleq \\
\Scale[0.8]{
\textbf{J}_{dB}(\bm{X}^w)     
\begin{bmatrix}
\bm{\Sigma}_{\beta}(\bm{p}_i)          
& \bm{0} \\
\bm{0} & 
\bm{\Sigma}_{\beta}(\bm{q}_i)            
\end{bmatrix} 
\textbf{J}_{dB}(\bm{X}^w)^T
}    
\end{equation}
\begin{equation}\label{Eq:Jd3D}
\textbf{J}_{dB}(\bm{X}^w) \triangleq
\frac{\partial (d_{3D}(i, k,\bm{X}^w) + \mu d_{P}(\bm{x}_i, k,\bm{X}^w)) }{\partial (\bm{\beta}_k(\bm{p}_i),\bm{\beta}_k(\bm{q}_i))}\in \mathbb{R}^{1 \times 6}
\end{equation}
\begin{equation}\label{Eq:SigmaBeta1}
\bm{\Sigma}_{\beta}(\bm{x}) \triangleq
\Scale[0.9]{\frac{\partial \bm{\beta}_k(\bm{x})}{\partial (u,v,\delta)} }         
\Scale[0.9]{\begin{bmatrix}
       \sigma_{li}^2 & 0 & 0 \\
       0 & \sigma_{li}^2 & 0 \\       
       0 & 0 & \sigma_{z}^2(\delta)  
     \end{bmatrix}} \Scale[0.9]{\frac{\partial \bm{\beta}_k(\bm{x})}{\partial (u,v,\delta)} ^T} 
     \in \mathbb{R}^{3 \times 3}
\end{equation}
where  $\textbf{J}_{dB}(\bm{X}^w) \in \mathbb{R}^{1 \times 6}$ is the Jacobian of eq.~(\ref{Eq:3DDistancePointLineSegment}) with respect to the coordinates of both the backprojected endpoints $\bm{\beta}_k(\bm{p}_i)$ and $\bm{\beta}_k(\bm{q}_i)$, $\frac{\partial \bm{\beta}_k(\bm{x})}{\partial (u,v,\delta)} \in \mathbb{R}^{3 \times 3}$ is the Jacobian of eq.~(\ref{Eq:Backprojection}) with respect to $u,v,\delta$. It is worth noting that \textit{(i)} for simplicity, we imposed $\bm{\Sigma}_{Li}$ is diagonal  \textit{(ii)} $\bm{\Sigma}_{\beta}(\bm{x}_i)$ represents the (induced) noise covariance matrix of the 3D backprojected point $\bm{\beta}_k(\bm{x})$.

For a \textit{stereo} line observation, we use the following 3D \emph{backprojection error}:
\begin{align}\label{Eq:3DBackprojectionError}
e_{3D}^L(i, k) &\triangleq   \rho \bigg( \|
\bm{d}_B(i, k) \|_{\bm{\Sigma}_{Li}}  \bigg)
\end{align}
that combines the 3D line-line distances and the 3D endpoint-endpoint distances as illustrated in Fig.~\ref{Fig:LineErrors}. 

To properly compute the distances $d_P$ (which are combined in the backprojection error~(\ref{Eq:3DBackprojectionError})), we need to correctly associate $\bm{\beta}_k(\bm{p}_i)$ with $\bm{P}^w$ and $\bm{\beta}_k(\bm{q}_i)$ with $\bm{Q}^w$. This is achieved when BA is initialized by computing the distances between the two candidate pairings $(\bm{\beta}_k(\bm{p}_i),\bm{P}^c)$ and $(\bm{\beta}_k(\bm{p}_i),\bm{Q}^c)$, and then actually the closest pairs of points.

In eq.~(\ref{Eq:3DDistanceBB}), the second terms $d_P$ act as a soft constraint (or regularization term) which naturally constrains the parametrization (the endpoints do not run away along lines) and allows to obtain better-configured endpoints. Indeed, the minimization of the 3D backprojection error~(\ref{Eq:3DBackprojectionError}) has two positive effects: \textit{(i)} In each camera frame, the 3D line segment endpoints $(\bm{P},\bm{Q})$ are aligned to their corresponding backprojected counterparts $(\bm{\beta}_k(\bm{p}_i),\bm{\beta}_k(\bm{q}_i))$ that convey available depth information (cfr. eq~(\ref{Eq:3DDistancePointLineSegment})) \textit{(ii)} each 3D map endpoint $\bm{P}$ $(\bm{Q}$) is constrained to stay close as much as possible to the centroid of all its backprojected representations $\bm{\beta}_k(\bm{p}_i)$ ($\bm{\beta}_k(\bm{q}_i)$) and does not run away (cfr. eq~(\ref{Eq:3DDistancePointBackProjPoint}) and see figures \ref{Fig:LineRepresentation} and~\ref{Fig:LineErrors}).

\begin{figure}[!t]
\begin{center}
{\centering\includegraphics[scale=0.25]{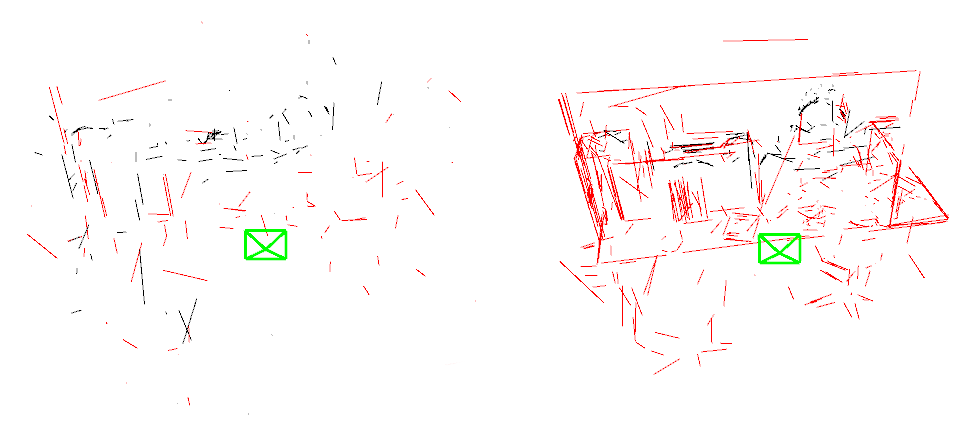}} \caption{Two 3D maps of line segments built for the same enviroment. \emph{Left}: the map built without using the backprojection error. \emph{Right}: map obtained by including the backprojection error in BA.} \label{Fig:MapsWithoutWithBackprojErrors}
\end{center}
\end{figure}

In practice, many experiments confirmed that without the usage of the 3D backprojection errors, the positions of the 3D line segment endpoints $\bm{P}^w_j, \bm{Q}^w_j$ usually drift along the line directions or the estimate of the line directions tends to degrade. In most cases, the line-segment-culling removes many unstable line segments from ${\cal M}_s$ after a couple of frames, since these do not succeed in being matched and tracked in the subsequent frames. 
Fig.~\ref{Fig:MapsWithoutWithBackprojErrors} shows two maps of the same environment obtained with and without including the backprojection error in BA. 

In the appendix, we detail how to compute the differentials of the reprojection and backprojection errors along with their covariance matrices. 

\subsection{Sparse Map Bundle Adjustment}

The back-end of the sparse SLAM uses \emph{full} Bundle Adjustment (BA) to compute  an optimal estimate of the model parameters in the sparse map ${\cal M}_s$. These are the poses of the keyframes $\cal K$, the positions of the 3D feature points  $\cal P$, and the 3D endpoint positions of the line segments $\cal L$. On the other hand, \emph{local} BA optimizes only the model parameters of points, line segments, and poses corresponding to a set of keyframes that are covisible with the current reference keyframe~\cite{mur2017orb}. 

In BA, the optimality of the estimates is implicitly defined by the reprojection errors which are minimized. Under a suitable mathematical formulation, BA is equivalent to a MAP or MLE estimation depending on the presence of a prior~\cite{cadena2016past, hartley2003multiple}. 

The complete objective function that is minimized by full BA is
\begin{align}\label{Eq:FullObjectiveFunInBA}
E &\triangleq \underset{k \in {\cal K}}{\sum} \Bigg( \underset{i \in {\cal P}^m_k}{\sum} e_{2D}^P(i, k) + \underset{h \in {\cal P}^s_k}{\sum} e_{3D}^P(h, k) +  \nonumber \\ 
&\quad \underset{j \in {\cal L}^m_k}{\sum} e^L_{2D}(j, k)  + \underset{r \in {\cal L}^s_k}{\sum} \Big( e^L_{2D}(r, k)  + e^L_{3D}(r, k) \Big) \Bigg)
\end{align}
where ${\cal K}$ is the set of keyframes, ${\cal P}^x_k \subset {\cal P}$ and ${\cal L}^x_k \subset {\cal L}$ are respectively the subsets of points and line segments which are observed  in the $k$-th keyframe, and the superscript $x \in \{m,s\}$ denotes mono or stereo observations.  

The optimization is performed by using the Levenberg-Marquardt method, where the augmented normal equations 
\begin{equation}
(\textbf{J}^T \bm{\Sigma}^{-1} \textbf{J} + \lambda \textbf{I}) \Delta \bm{\theta} =  -\textbf{J}^T\bm{\Sigma}^{-1} \bm{\epsilon}
\end{equation}
are iteratively solved to obtain the update of the full vector $\bm{\theta}$ of map parameters. Here, the error vector $\bm{\epsilon}$ combines the reprojection errors and backprojection distances considered in eq.~(\ref{Eq:FullObjectiveFunInBA}), and $\bm{\Sigma}$ combines all the corresponding covariance matrices \cite{hartley2003multiple}.

%% file: sections/points.tex
\subsection{Point Reprojection Errors}\label{Sect:PointReprojectionErrors}

In this section, we describe the different point reprojection errors we used in PLVS.

Consider the generic feature point $p \in {\cal P}$ with its $n_p \in \mathbb{N}$ observations $\{ \bm{p}_i \}_{i=1}^{n_p}$ and its  Euclidean representation $\bm{P} \in \mathbb{R}^3$. In this regard, a point observation $\bm{p}_i= [u_i,v_i]^T \in \Omega$ is collected in a specific frame/keyframe $k$ and is considered \emph{stereo} if its corresponding depth $\delta_i=D_k(\bm{p}_i)$ is finite and exists; it is \emph{mono} otherwise.

The 2D reprojection error for a mono point observation is defined as 
\begin{equation}
e_{2D}^P(i, k) \triangleq  \rho \bigg( \Big\| \bm{p}_i -  \bm{\pi}_k(\bm{P}^w) \Big\|_{\Sigma_{Pi}}  \bigg)
\end{equation}
where the covariance matrix is $\Sigma_{Pi} = \sigma_{pi}^2 \textbf{I}_2$, $\sigma_{pi}^2 \in \mathbb{R}$ is the point detection noise variance\footnote{Feature points are extracted on different scales of a Gaussian image pyramid. 
The noise variance $\sigma^2_{pi}$ depends on the level on which $\bm{p}_i$ is detected.}, and $\rho$ is the robust Huber cost function~\cite{hartley2003multiple, huber2011robust}. 

In the stereo case, we assume the images are undistorted and rectified. With a binocular camera, a stereo point observation is $\bm{p}_i^s = [u^l_i,v^l_i,u^r_i]^T \in \mathbb{R}^3$ where the first two coordinates are the pixel measurements in the left image, and $u^r$ is the row measurement in the right frame. In such a case, the classic stereo reprojection error~\cite{strasdat2011double} is:
\begin{equation}\label{Eq:StereoPointProjError}
e_{3D}^P(i, k) \triangleq  \rho \bigg( \Big\| \begin{bmatrix} u^l_i \\ v^l_i \\ u^r_i \end{bmatrix} -  \begin{bmatrix} \bm{\pi}_k(\bm{P}^w) \\ f_x \frac{(x_c -b)}{z_c} + c_x \end{bmatrix} \Big\|_{\Sigma_{Si}}  \bigg)
\end{equation}
where $b$ is the stereo baseline, the covariance matrix is $\Sigma_{Si} = \sigma_{pi}^2 \textbf{I}_3$, and $\bm{P}^c = [x_c,y_c,z_c]^T = \textbf{R}_k \bm{P}^w + \bm{t}_k $. Here, we estimate the depth from the disparity as $\delta_i = b f_x/(u^l_i-u^r_i)$.

With an RGB-D camera, $u^r_i$ does not exist. However, it can be simulated by using a "virtual" baseline $b$ (a constant parameter) and computed as $u_i - \frac{b f_x}{\delta_i}$, where $\delta_i = D_k(\bm{p}_i)$ is the measured point depth. In this case, the stereo reprojection error is defined as: 
\begin{equation}\label{Eq:RGBDPointProjError}
e_{3D}^P(i, k) \triangleq  \rho \bigg( \Big\| \begin{bmatrix} \bm{p}_i \\ u_i - \frac{b f_x}{\delta_i} \end{bmatrix} -  \begin{bmatrix} \bm{\pi}_k(\bm{P}^w) \\ f_x \frac{(x_c -b)}{z_c} + c_x \end{bmatrix} \Big\|_{\Sigma_{Ri}}  \bigg)
\end{equation} 
where the covariance matrix is:
\begin{align}\label{Eq:StereoPointProjErrorCovariance}
\Sigma_{Ri} &= \textbf{J}_S \Scale[0.9]{\begin{bmatrix}
       \sigma_{pi}^2 & 0 & 0 \\
       0 & \sigma_{pi}^2 & 0 \\       
       0 & 0 & \sigma_{z}^2(d_i)  
     \end{bmatrix}} \textbf{J}_S ^T \\
     \quad &=
     \Scale[0.9]{ \begin{bmatrix}
       \sigma_{pi}^2 & 0 & \sigma_{pi}^2 \\
       0 & \sigma_{pi}^2  & 0 \\
      \sigma_{pi}^2   & 0 & \sigma_{pi}^2 + \frac{b^2 f_x^2}{\delta_i^4} \sigma_z(\delta_i)^2 
     \end{bmatrix}}
\end{align}
and $\textbf{J}_S = \frac{\partial}{\partial (u_i,v_i,\delta_i)}\begin{bmatrix} \bm{p}_i \\ u_i - \frac{b f_x}{\delta_i} \end{bmatrix} $ is the Jacobian of the norm argument in eq.~(\ref{Eq:RGBDPointProjError})  and $\sigma_{z}^2(\delta)$ is the axial depth noise model presented in~\cite{nguyen2012modeling}. With RGB-D cameras, we considered and tested both the covariance matrix~(\ref{Eq:StereoPointProjErrorCovariance}) and the simpler covariance $\Sigma_{Si} = \sigma_{pi}^2  \bm{I}_3$, which is applied with stereo rigs. In principle, $\Sigma_{Si}$ does not take into account \textit{(i)} the correlation between the first and the third components of the stereo reprojection error~(\ref{Eq:StereoPointProjError}) and \textit{(ii)} the actual relationship between the axial depth uncertainty $\sigma_{z}^2$ and the measured depth $\delta_i$ (as described in~\cite{nguyen2012modeling}). However, note that the calibration procedure may introduce modeling errors and does not necessarily bring better results. 

As a further alternative, we also consider the following stereo reprojection error which is specific to RGB-D cameras
\begin{equation}\label{Eq:StereoPointProjErrorDepth}
e_{3D}^P(i, k) \triangleq  \rho_{\omega_3} \bigg( \Big\| \begin{bmatrix} \bm{p}^s_i \\ \delta_i \end{bmatrix} -  \begin{bmatrix} \bm{\pi}_k(\bm{P}^w) \\ z_c \end{bmatrix} \Big\|_{\Sigma_{Si}}  \bigg)
\end{equation}
and comes with a simpler diagonal covariance matrix:
\begin{equation}
\Sigma_{Si} = 
      \begin{bmatrix}
       \sigma_{pi}^2 & 0 & 0 \\
       0 & \sigma_{pi}^2  & 0 \\
      0   & 0 & \sigma_z(\delta_i)^2 
     \end{bmatrix}.
\end{equation}

In our tests, we observed that the simple stereo covariance $\Sigma_{Si} = \sigma_{pi}^2  \bm{I}_3$ allows to get more stable and accurate results.

%% file: sections/voma.tex
Each new keyframe generated by the SLAM module enters as input in the VOMA. As a first step, the VOMA backprojects the depthmap $D_i$ of each new keyframe into a point cloud, represented in the camera frame. To this aim, we use the map $\bm{\beta}_i(\bm{u}) = \textbf{K}^{-1}  \widebar{\bm{u}} D_i(\bm{u})$ over camera pixels~(see Sect.~\ref{Sect:Notation}). 

In parallel, normals are incrementally and robustly computed on each RGB-D image at backprojection time. In particular, on each 3D point of the current backprojected depth image, we average (by area weighting \cite{klasing2009comparison}) the 4 normals obtained by central differencing on a 4-connected neighborhood. A result of our normals computation is shown in Fig.~\ref{Fig:PLVSDetail}b. Normals coming from different RGB-D images are further integrated and averaged in our custom octree-based map (see the following subsection~\ref{Sect:OctreeBasedMap}).

Different methods are available to fuse the computed keyframe point clouds. In particular, octrees, octomap, and spatially-hashed voxels with Truncated Signed Distance
Fields (TSDFs) and meshes. We briefly present them in the following subsections.

\subsection{Octree-based Map}\label{Sect:OctreeBasedMap}

In this case, we use an octree data structure to efficiently represent a 3D  volumetric map. Two different octree-based volumetric maps are available.
\begin{enumerate}
\item Our custom octree-based model: Each voxel represents a centroid that averages all the integrated points (coordinates, colors, and normals). Projective association and space carving can be enabled to cope with dynamic changes occurring in the environment.
\item The fastfusion method~\cite{Steinbruecker2014}: This is a very efficient multi-scale octree that stores a Signed Distance Function (SDF) at multiple levels.
\end{enumerate}
 
\subsection{Octomap Model}

In this well-known approach \cite{hornung13auro}, a 3D occupancy grid is used to model the 3D space volume. New point cloud integrations are implemented in a probabilistic fashion by using suitable sensor models and raycasting operations. Colors of points belonging to the same voxel are averaged.

\subsection{Spatially-hashed Voxels with TSDF}

Spatially-hashed data structures with TSDF are used to represent the 3D space volume~\cite{curless1996volumetric, newcombe2011kinectfusion}. In general, TSDFs are fast to build and smooth out sensor noise over many observations. In this context, a mesh model is extracted from the built TSDF to represent the surrounding 3D surfaces by using the Marching Cubes method~\cite{lorensen1987marching}. Two different TSDF-based approaches are available.  

\begin{enumerate}
\item Chisel~\cite{klingensmith2015chisel}: we tweaked and improved some portions of the original open-source code to attain better real-time performances. 
\item Voxblox~\cite{oleynikova2017voxblox}: this approach provides an efficient implementation and was originally proposed to incrementally build Euclidean Signed Distance Fields (ESDFs) out of TSDFs in dynamically growing maps (for planning purposes).
\end{enumerate}

\subsection{Implementation Details}

In the VOMA, a FIFO queue collects all the new keyframes that arrive from the SLAM module. Keyframes are processed and integrated $N$ at a time in the volumetric map model, where $N$ is an integer parameter defined by the user. Subsequently, a volumetric model update is performed. The latter consists of a possible mesh model update, depending on the adopted volumetric method (for instance, by using incremental Marching Cubes), and the extraction of a global point cloud map (which is used for instance by the visualizer or made available for other purposes).   

In general, the update of the mesh model and the zero level extraction from the TSDF (e.g., by using Marching Cubes) take more time than a single point cloud integration. The reason is that the mesh model update needs to take into account more information coming from voxel neighborhood relations.

To maintain consistency between the sparse and volumetric maps, the latter is rebuilt (or adjusted) whenever the SLAM map or its underlying pose graph goes through a global adjustment (e.g. when loop closures are detected).  

%% file: sections/segmentation.tex
Our incremental segmentation approach consists of an adapted version of the method presented in \cite{tateno2015real}. It is available only with RBG-D cameras and when our custom octree-based VOMA is active. With respect to the original approach, during the depth image segmentation stage, we also integrate available line segment information to better define the segment borders. We refer the Reader to the mentioned paper~\cite{tateno2015real} for further design and implementation details.


%% file: sections/exp-results-summary.tex
\begin{figure*}[ht]
\centering
\subfigure[]{\includegraphics[height=6cm]{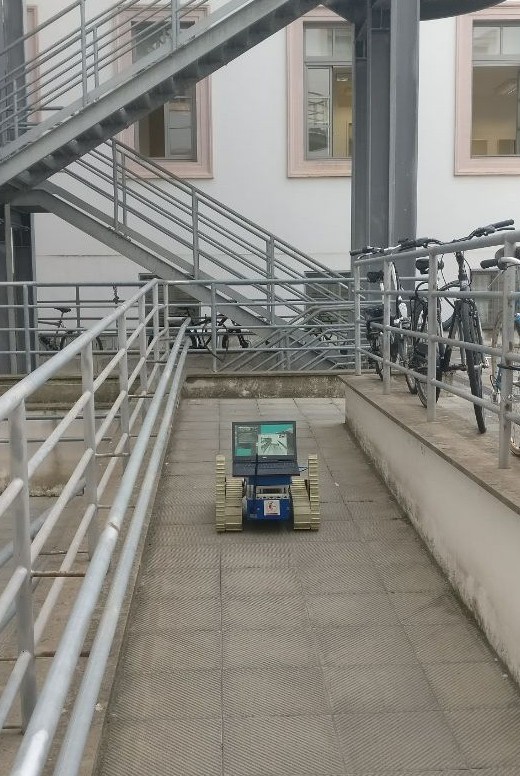}}\quad
\subfigure[]{\includegraphics[height=6cm]{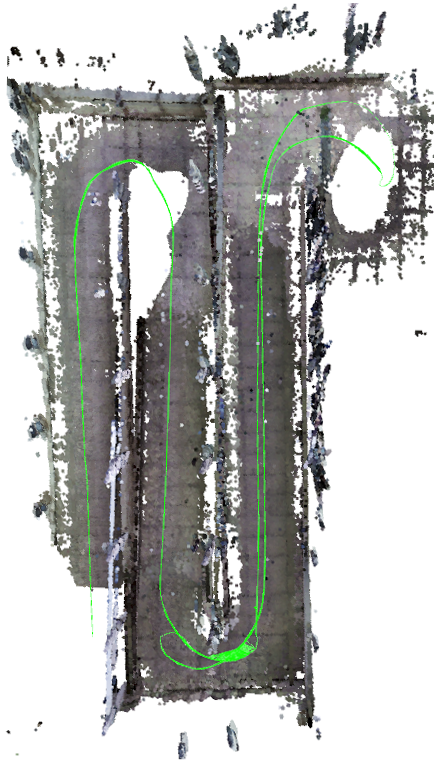}}\quad
\subfigure[]{\includegraphics[height=6cm]{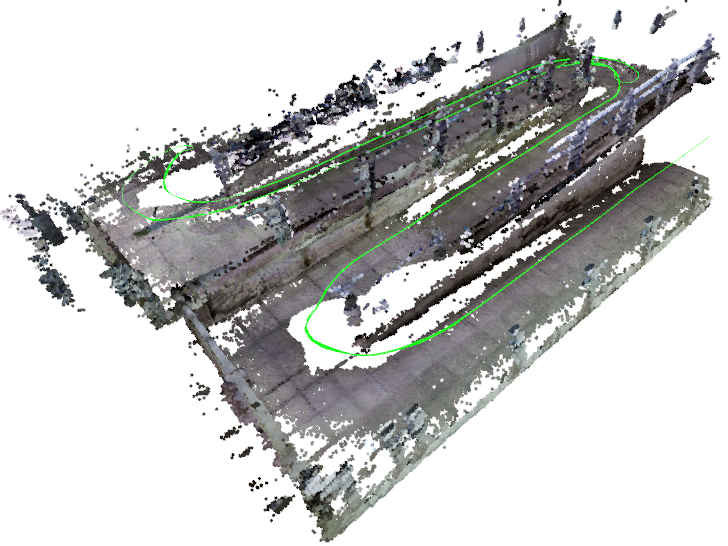}}
\caption{The robot equipped with a realsense R200 and the built volumetric map (based on the octree model described in Sect.).}
\label{Fig:RealsenseMap}
\end{figure*}

\begin{figure*}[!t]
\centering
\subfigure[]{\includegraphics[scale=0.3]{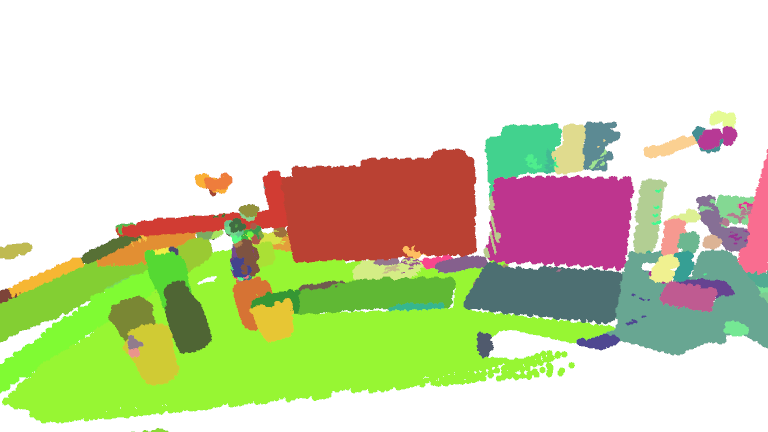}}
\subfigure[]{\includegraphics[scale=0.35]{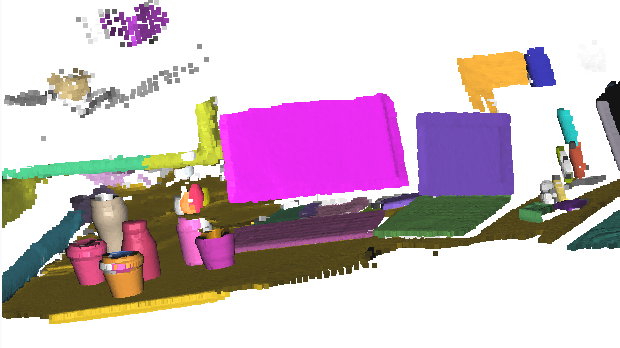}}
\caption{Incremental segmentation detail (a) PLVS (b) inSeg~\cite{tateno2015real}}
\label{Fig:segAcc}
\end{figure*}

\begin{figure*}[!t]
\centering
\subfigure[]{\includegraphics[scale=0.17]{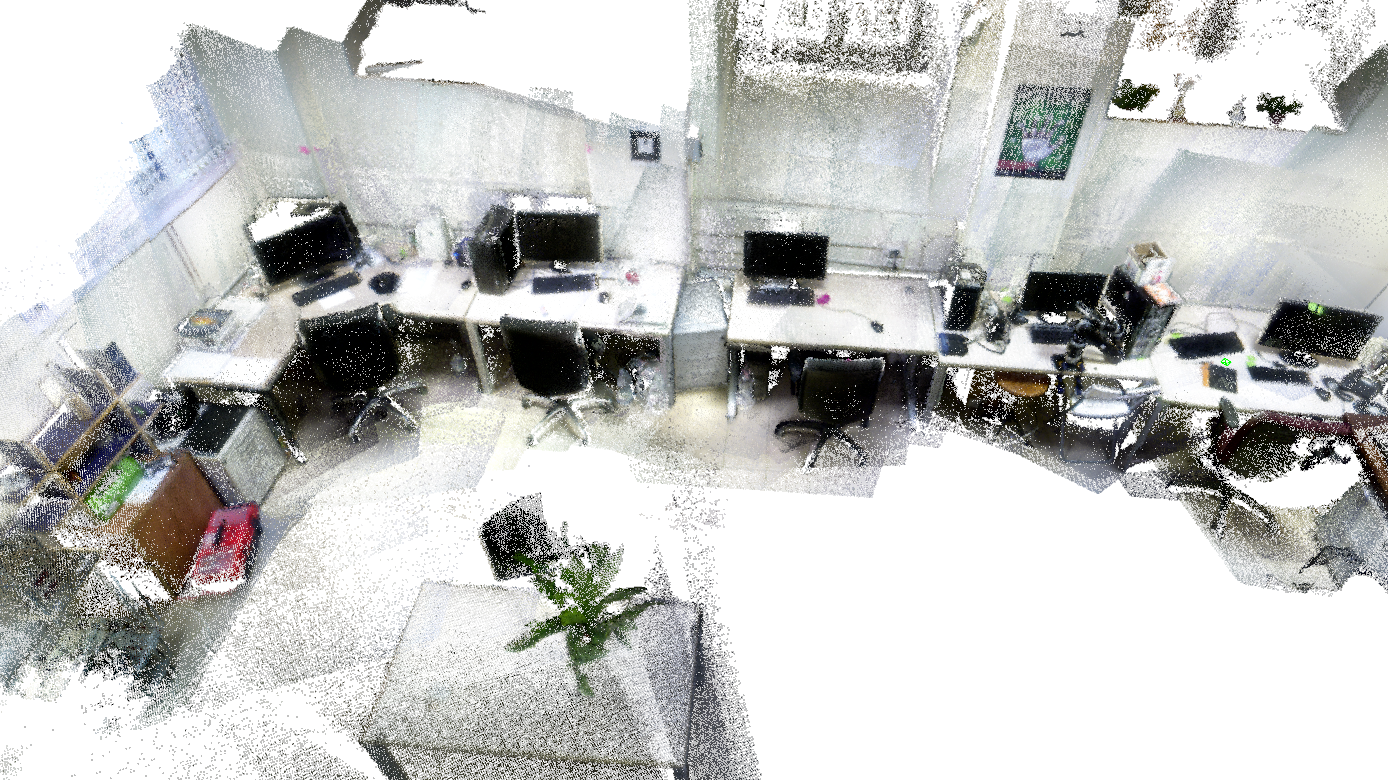}}
\subfigure[]{\includegraphics[scale=0.17]{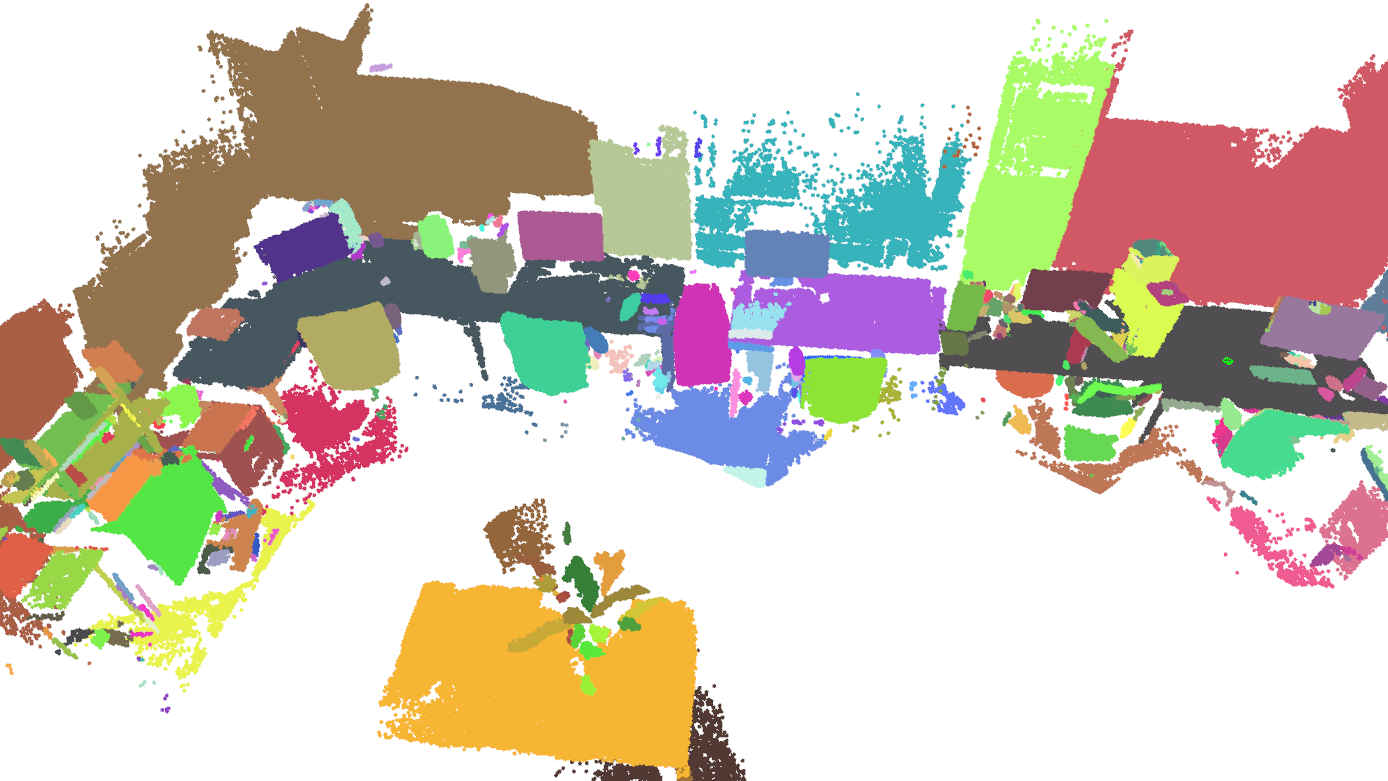}}
\caption{Segmentation performed on big office environment}
\label{Fig:bigSeg}
\end{figure*}

We successfully tested the PLVS framework live with different sensors (Asus Xtion Pro, ZED stereo camera, Intel Realsense, etc.) to prove its versatility. Figure~\ref{Fig:RealsenseMap} shows a volumetric map (custom octree-based method) obtained with a tracked robot equipped with a realsense R200. Figure~\ref{Fig:bigSeg} shows the 3D reconstruction and segmentation of a laboratory we obtained live with a hand-held Asus Xtion~Pro. 

The results of a first qualitative and quantitative evaluation of the PLVS I framework are reported in a separate document~\cite{fredaPLVSeval2023}, which is online available. The obtained evaluation results show that PLVS I has localization and 3D reconstruction accuracies that are on par with systems that are well-known in the SLAM literature. Notably, we used the same set of PLVS parameters on all the TUM RGB-D sequences (no fine-tuning per dataset). 

The PLVS project webpage provides videos and further system design details~\cite{fredaPLVSweb}. 
An open-source implementation is available at \href{https://github.com/luigifreda/plvs}{github.com/luigifreda/plvs}.

%% file: sections/conclusions.tex
We presented PLVS, a modular and versatile system that shows the tracking "agility" of feature-based SLAM systems and is able to generate dense volumetric maps solely relying on CPU.
Different volumetric mapping methods are available: octrees, octomap, and spatially-hashed voxels with TSDFs and meshes. The different capabilities of the system are organized into divisions that can be enabled/disabled and configured by users in different ways. This allows to finely trade-off map accuracy/resolution versus CPU load and to adapt the framework capabilities to the system at hand. 

The results of a first qualitative and quantitative evaluation of the PLVS~I framework are reported in a separate document~\cite{fredaPLVSeval2023}, which is online available. The obtained evaluation results show that PLVS~I has localization and 3D reconstruction accuracies that are on par with systems that are well-known in the SLAM literature. Notably, we used the same set of PLVS parameters on all the evaluated sequences (no fine-tuning per dataset).

PLVS is an active project. The system has evolved under many aspects from the date (years ago) in which we performed our first evaluation. Now, PLVS is available in two different versions and has integrated many new improvements over time. The first version, PLVS I, is based on ORB-SLAM2, while the newest one, PLVS II, is based on ORB-SLAM3 and supports camera systems equipped with IMUs.  We are working on a new evaluation of both PLVS I and PLVS II systems. New results will be shared soon. The PLVS project webpage provides videos and further information~\cite{fredaPLVSweb}.

%% file: sections/appendix.tex
\subsection{Stereo Line Triangulation with Closed-form Solution} \label{Sect:LineTriangulation}

Consider two camera poses\footnote{In this section, for convenience, we override some of the symbols used beforehand. The aim is to reflect the standard notation used in well known computer vision books~\cite{hartley2003multiple}.}. Let $\textbf{P}_1=\textbf{K}_1 [ \textbf{R}_1 \ \bm{t}_1 ]$  and $\textbf{P}_2=\textbf{K}_2 [ \textbf{R}_2 \ \bm{t}_2 ]$ be their corresponding camera projection matrices. In particular, $\textbf{R}_i=\textbf{R}_{c_iw} \in SO(3)$,  and $\bm{t}_i=\bm{t}_{c_iw} \in \mathbb{R}^3$ denote the rotation and translation (respectively) of the transformation $\textbf{T}_{c_iw} \in SE(3)$ from the world frame to the $i$-th camera frame (i.e. $\bm{X}^{c_i} = \textbf{R}_{c_iw} \bm{X}^w + \bm{t}_{c_iw}$).

Refer to Fig.~\ref{Fig:LineRepresentation} and let $\bm{p}_i = [p_u, p_v, 1]^T$ and $ \bm{q}_i=[q_u, q_v, 1]^T$ be homogeneous representations of the 2D line segment endpoints on image $i$. The 2D image lines $\bm{l}_1 = \bm{p}_1 \times \bm{q}_1$ and $\bm{l}_2 = \bm{p}_2 \times \bm{q}_2$ \mbox{back-project} to the 3D space planes $\bm{\alpha}_1 = \textbf{P}_1^T \bm{l}_1 \in \mathbb{R}^4$ and $\bm{\alpha}_2 = \textbf{P}_2^T \bm{l}_2 \in \mathbb{R}^4$ respectively~\cite{hartley2003multiple}. In fact, a 3D space point $\bm{X}^w \in \mathbb{R}^3$ is projected on the 2D image line $\bm{l}_i$ if and only if $\bm{l}^T (\bm{P}_i\widebar{\bm{X}}^w)=0$: this implies $\bm{\alpha}_i^T\widebar{\bm{X}}^w=0$.  

Our goal is to compute the space line segment at the intersection of the two planes $\bm{\alpha}_1$ and $\bm{\alpha}_2$. To this end, we compute the intersections between the two rays \mbox{back-projected} from the endpoints $\bm{p}_1$ and $\bm{q}_1$, and the plane $\bm{\alpha}_2$. 

Let $\bm{x}$ be a generic point on image 1 that \mbox{back-projects} to the 3D space point $\bm{X}^c = \lambda \textbf{K}_1^{-1} \bm{p}_1$ that lies on $\bm{\alpha}_2$, where $\lambda \in \mathbb{R}^+$ is the unknown depth. In the world frame, one has:
\begin{equation}
\bm{X}^w = \lambda \textbf{R}_{wc_1} \textbf{K}_1^{-1} \bm{p}_1 +  \bm{t}_{wc_1} = \lambda \textbf{R}_1^T \textbf{K}_1^{-1} \bm{p}_1 -  \textbf{R}_1^T \bm{t}_1
\end{equation}
where $\textbf{R}_{wc_1}=\textbf{R}_{c_1w}^T=\textbf{R}_1^T$ and $\bm{t}_{wc_1} = -\textbf{R}_{c_1w}^T \bm{t}_{c_1w}=-\textbf{R}_1^T \bm{t}_1$. Since $\bm{X}^w$ lies on  $\bm{\alpha}_2$ then $\bm{\alpha}_2^T\widebar{\bm{X}}^w=0$, that is:
\begin{equation}
\bm{l}_2^T \textbf{K}_2 [ \textbf{R}_2 \ \bm{t}_2 ] \begin{bmatrix}
       \lambda \textbf{R}_1^T \textbf{K}_1^{-1} \bm{p}_1 -  \textbf{R}_1^T \bm{t}_1 \\
       1  
\end{bmatrix} = 0.
\end{equation}
This implies: 
\begin{equation}
\lambda \bm{l}_2^T \textbf{K}_2 \textbf{R}_2 \textbf{R}_1^T \textbf{K}_1^{-1} \bm{p}_1 =  \bm{l}_2^T \textbf{K}_2 (\textbf{R}_2 \textbf{R}_1^T \bm{t}_1 - \bm{t}_2)
\end{equation}
and we obtain: 
\begin{equation}
\lambda  =  \frac{\bm{l}_2^T \textbf{K}_2 (\textbf{R}_2 \textbf{R}_1^T \bm{t}_1 - \bm{t}_2)}{\bm{l}_2^T \textbf{K}_2 \textbf{R}_2 \textbf{R}_1^T \textbf{K}_1^{-1} \bm{p}_1}.
\end{equation}

This last equation can be simplified by using the two following identities: 
\begin{equation}
\textbf{R}_{c_2c_1} = \textbf{R}_{c_2w} \textbf{R}_{wc_1} =  \textbf{R}_2 \textbf{R}_1^T \triangleq  \textbf{R}_{21} 
\end{equation}
\begin{multline}
\bm{t}_{c_2c_1} = \textbf{R}_{c_2w}(\bm{t}_{wc_1}-\bm{t}_{wc_2}) = \textbf{R}_2(-\textbf{R}_1^T \bm{t}_1 + \textbf{R}_2^T\bm{t}_2) = \\
-\textbf{R}_2 \textbf{R}_1^T \bm{t}_1 + \bm{t}_2 \triangleq \bm{t}_{21}
\end{multline}
where $\textbf{R}_{c_2c_1}$ and $\bm{t}_{c_2c_1}$ denote the rotation and translation (respectively) of the transformation $\textbf{T}_{c_2c_1} \in SE(3)$ from camera 1 to the camera 2.
In particular, we obtain:  
\begin{equation}\label{Eq:DepthLineTriangulation}
\lambda  =  \frac{-\bm{l}_2^T \textbf{K}_2 \bm{t}_{21}}{\bm{l}_2^T \textbf{K}_2 \textbf{R}_{21} \textbf{K}_1^{-1} \bm{x}} =  \frac{-\bm{l}_2^T \bm{e}_2 }{\bm{l}_2^T \textbf{H}_{21} \bm{x}}
\end{equation}
where $\bm{e}_2 = \textbf{K}_2 \bm{t}_{21}$ is the epipole in image 2 and $\textbf{H}_{21} = \textbf{K}_2 \textbf{R}_{21} \textbf{K}_1^{-1}$ is the infinite homography mapping vanishing points from camera 1 to camera 2~\cite{hartley2003multiple}. Finally, we can use eq.~(\ref{Eq:DepthLineTriangulation}) to estimate the depths of the line segment endpoints $\bm{p}_1$ and $\bm{q}_1$ of $\bm{l}_1$. Note that the triangulation becomes unstable close to the degenerate configuration $\bm{l}_2^T \textbf{H}_{21} \bm{x}=0$: that is, when on image 2 the vanishing point corresponding to $\bm{x}$ lies on $\bm{l}_2$ (in other words, $\bm{\alpha}_2$ is parallel to the epipolar plane of $\bm{x}$~\cite{liu20233d}). In such a degenerate case: \textit{(i)} if $\bm{l}_2^T \bm{e}_2 = 0$ (the epipole $\bm{e}_2$ belongs to $\bm{l}_2$) there are infinite solutions since $\bm{l}_2 = \bm{e}_2 \times \textbf{H}_{21} \bm{x}$ is an epipolar line and $\bm{\alpha}_2$ coincides with the epipolar plane of $\bm{x}$; \textit{(ii)} otherwise, there are no solutions.

In the special case of an ideal binocular camera (i.e. with undistorted and rectified stereo images), one has $\textbf{R}_{21}=\textbf{I}$, $\textbf{K}_{2}=\textbf{K}_{1}$  and $\bm{t}_{21} =[-b,0,0]^T$ where $b \in \mathbb{R}^+$ is the baseline. From eq.~(\ref{Eq:DepthLineTriangulation}), we obtain: 
\begin{equation}
\lambda  =  \frac{b f_x l_2^x}{\bm{l}_2^T\bm{x}} \qquad  d = \frac{\bm{l}_2^T\bm{x}}{l_2^x}
\end{equation}
where $\bm{l}_2 = [l_2^x, l_2^y, l_2^z]^T$ and $d \in \mathbb{R}$ is the corresponding stereo disparity. In this case, the epipolar lines are horizontal and the degenerate case occur when the 2D image lines $\bm{l}_1$ and $\bm{l}_2$ are parallel (hence $\bm{l}_2^T\bm{x}=0$). 

\subsection{Useful Derivative Results}

In the remainder of this  appendix, we will use the following results. Let $\bm{v}(t)$ be a vector  in $\mathbb{R}^2$ or $\mathbb{R}^3$ which is a function of $t$. Denote by $\bm{n}(t) = \frac{\bm{v} }{\| \bm{v}\|}$ its corresponding unit vector. One has 
\begin{align}
\label{Eq:ModulusDerivative}
\frac{\partial \Vert \bm{v}(t) \Vert}{\partial t} &= \frac{\partial \sqrt{\bm{v}^T\bm{v}}}{\partial t} = \frac{ \bm{v}^T}{\Vert \bm{v} \Vert} \frac{\partial  \bm{v}}{\partial t} \\ 
\label{Eq:InvModulusDerivative}
\frac{\partial }{\partial t}\frac{1}{\|\bm{v}(t)\|}  &= \frac{\partial }{\partial t}\frac{1}{\sqrt{\bm{v}^T\bm{v}}} = -\frac{\bm{v}^T}{\|\bm{v}\|^3}\frac{\partial \bm{v}}{\partial t}
\\
\frac{\partial \bm{n}(t)}{\partial t} &= \frac{\partial }{\partial t}\frac{\bm{v}}{\|\bm{v}\|} = \frac{1}{\|\bm{v}\|}\frac{\partial \bm{v}}{\partial t} + \bm{v} \frac{\partial }{\partial t}\Big( \frac{1}{\|\bm{v}\|} \Big) \nonumber \\ 
\label{Eq:NormalDerivative}
&= \frac{1}{\|\bm{v}\|} \Big( \textbf{I} - \bm{n} \bm{n}^T \Big)\frac{\partial \bm{v}}{\partial t}
\end{align}
Deriving a cross product in $\mathbb{R}^3$, it is 
\begin{align}
\label{Eq:CrossProductDerivative}
\frac{d (\bm{v} \times \bm{w})}{dt} &= \frac{d\bm{v}}{dt}\times \bm{w} + \bm{v} \times \frac{d\bm{w}}{dt} =  \widehat{\bm{v}}\frac{d\bm{w}}{dt} -\widehat{\bm{w}}\frac{d\bm{v}}{dt}
\end{align}
where $\widehat{\bm{x}} \in \mathbb{R}^{3 \times 3}$ denotes the skew-symmetric matrix corresponding to $\bm{x}=[x,y,z]^T \in \mathbb{R}^3$~\cite{barfoot2017state, ma2012invitation}. In particular:
\begin{equation}
\widehat{\bm{x}} = \begin{bmatrix}
       0 & -z & y \\
       z & 0  & -x \\
       -y & x  & 0
\end{bmatrix} \in \mathbb{R}^{3 \times 3} \qquad \bm{x} \in \mathbb{R}^3.
\end{equation}
This corresponds to the classic definition of the hat operator $\widehat{(\cdot)}$ in $\mathbb{R}^3$
Notably, it is: $\bm{x} \times \bm{y} = \widehat{\bm{x}} \bm{y} = -\widehat{\bm{y}}\bm{x}$ with $\bm{x}, \bm{y} \in \mathbb{R}^3$.

\subsection{Exponential Map}

In BA optimization, we consider a \textit{left perturbation model} on $SE(3)$ in order to iteratively update at each step $k$ the estimate of a camera transformation matrix $\textbf{T} \in SE(3)$ in its manifold. Namely, let $\textbf{T}_k$ be the estimate of the transformation $\textbf{T}$ at step $k$. We apply left-multiplicative exponential increments in the form $\textbf{T}_{k+1} = e^{\widehat{\bm{\xi}}} \textbf{T}_k$, where $\widehat{\bm{\xi}} \in se(3)$ is the \emph{twist}, which belongs to the Lie-Algebra $se(3)$ and corresponds to the twist coordinates $\bm{\xi} \in \mathbb{R}^6$ \cite{barfoot2017state, ma2012invitation}. $\bm{\xi}$ is the variable we optimize to get $\textbf{T}_{k+1}$. The hat operator $\widehat{(\cdot)}$ is overladed in $\mathbb{R}^6$ as follows:
\begin{equation}
\widehat{\bm{\xi}} =  \widehat{
\begin{bmatrix}
       \bm{\phi}  \\
       \bm{\rho}  
\end{bmatrix}
} = 
\begin{bmatrix}
       \widehat{\bm{\phi}} & \bm{\rho} \\
       \bm{0}^T & 0  
\end{bmatrix} \in \mathbb{R}^{4 \times 4} \qquad \bm{\xi} \in \mathbb{R}^6, \quad \bm{\phi}, \bm{\rho} \in \mathbb{R}^3
\end{equation}
where $\bm{\phi} \in \mathbb{R}^3$ is the rotation vector and $\bm{\rho}$ represents a "translation part"\footnote{Note that $\bm{\rho}$ does not represent an actual translation vector in the transformation $e^{\widehat{\bm{\xi}}}$ (see also eq.~(\ref{Eq:expSE3})). When considering the $SE(3)$ differential equation $\dot{\textbf{T}}=\widehat{\bm{\xi}}\textbf{T}$, then $\bm{\xi}$ represents a \textit{generalized velocity} where $\bm{\phi}$ and $\bm{\rho}$ are the angular velocity and  the linear velocity respectively \cite{barfoot2017state, ma2012invitation}. In fact, let $\bm{x}_w, \bm{x}_c \in \mathbb{R}^3$ denote the coordinates of a 3D point in the \textit{world} and \textit{camera} frames respectively. Define $\bm{p}_w = [\bm{x}_w^T, 1]^T \in \mathbb{R}^4$ and $\bm{p}_c = [\bm{x}_c^T, 1]^T \in \mathbb{R}^4$. One has $ \dot{\bm{p}}_w = \dot{\textbf{T}}_{wc}\bm{p}_c=\widehat{\bm{\xi}}\textbf{T}_{wc}\bm{p}_c=\begin{bmatrix}
       \widehat{\bm{\phi}} & \bm{\rho} \\
       \bm{0}^T & 0  
\end{bmatrix} \bm{p}_w$. This entails $\dot{\bm{x}}_w = \widehat{\bm{\phi}} \bm{x}_w + \bm{\rho} = \bm{\phi} \times \bm{x}_w + \bm{\rho}$, which is a classic kinematics equation with $\bm{\phi}$ and $\bm{\rho}$ acting as angular velocity and linear velocity respectively.}. 
The matrix exponential is given by:
\begin{equation}
e^{\textbf{A}}=\overset{\infty}{\underset{k=0}{\sum}} \frac{1}{k!} \textbf{A}^k \in \mathbb{R}^{n \times n}, \qquad \textbf{A} \in \mathbb{R}^{n \times n}
\end{equation}
whose first order approximation is: 
\begin{equation}\label{Eq:ExpFirstOrderApprox}
e^{\textbf{A}} \approx \textbf{I} + \textbf{A}.
\end{equation}
In $SO(3)$, it is: 
\begin{equation}
\label{Eq:RodriguesEquation}
e^{\widehat{\bm{\phi}}}= \overset{\infty}{\underset{k=0}{\sum}} \frac{1}{k!} (\widehat{\bm{\phi}})^k = \cos\phi\textbf{I} + (1-\cos\phi )\bm{a}\bm{a}^T + \sin\phi\widehat{\bm{a}} 
\end{equation}
where $\phi=|\bm{\phi}|\in \mathbb{R}$ is the angle of rotation and $\bm{a}=\bm{\phi}/\phi$ is the unit-length axis of rotation. Eq.~(\ref{Eq:RodriguesEquation}) is the well known Rodrigues' formula. 

In $SE(3)$, one has:  
\begin{equation}
\label{Eq:expSE3}
e^{\widehat{\bm{\xi}}}=\overset{\infty}{\underset{k=0}{\sum}} \frac{1}{k!} (\widehat{\bm{\xi}})^k=\begin{bmatrix}
       e^{\widehat{\bm{\phi}}} & \textbf{J}\bm{\rho} \\
       \bm{0}^T & 1  
\end{bmatrix}, \qquad \textbf{J}=\overset{\infty}{\underset{k=0}{\sum}} \frac{1}{(k+1)!} (\widehat{\bm{\phi}})^k 
\end{equation}
where $\textbf{J} \in \mathbb{R}^{3 \times 3}$ has the closed form expression~\cite{barfoot2017state}:
\begin{equation}
\label{Eq:ClosedFormJ}
\textbf{J} = \frac{\sin\phi}{\phi}\textbf{I} + \big(1-\frac{\sin\phi}{\phi}\big)\bm{a}\bm{a}^T + \frac{1-\cos\phi}{\phi}\widehat{\bm{a}}.
\end{equation}

\subsection{Derivative on SE(3)}

It is possible to show that ~\cite{drummond2000application,sattinger2013lie}: 
\begin{equation}\label{eq:ExpDerivatives}
e^{\widehat{\bm{\xi}}}= e^{\overset{6}{\underset{j=1}{\sum}}  \textbf{G}_j \xi_j }, \quad \frac{d}{d\xi_j}e^{\widehat{\bm{\xi}}}\bigg|_{\bm{\xi}=0} =  \textbf{G}_j \quad j=1,...,6, \quad \bm{\xi} \in \mathbb{R}^6
\end{equation}
where the matrices $\textbf{G}_j \in \mathbb{R}^{4\times 4}$ are the selected $SE(3)$ \emph{group generators} and form a basis for $se(3)$:
\begin{align*}
\textbf{G}_1 &= \scaleG{\begin{bmatrix}
       0 & 0 & 0 & 0\\
       0 & 0 & -1 & 0\\
       0 & 1 & 0 & 0\\
       0 & 0 & 0 & 0
     \end{bmatrix}}\quad
\textbf{G}_2 = \scaleG{\begin{bmatrix}
       0 & 0 & 1 & 0\\
       0 & 0 & 0 & 0\\
       -1 & 0 & 0 & 0\\
       0 & 0 & 0 & 0
     \end{bmatrix}}\quad
\textbf{G}_3 = \scaleG{\begin{bmatrix}   
       0 & -1 & 0 & 0\\
       1 & 0 & 0 & 0\\
       0 & 0 & 0 & 0\\
       0 & 0 & 0 & 0
     \end{bmatrix}} \\
\textbf{G}_4 &= \scaleG{\begin{bmatrix}
       0 & 0 & 0 & 1\\
       0 & 0 & 0 & 0\\
       0 & 0 & 0 & 0\\
       0 & 0 & 0 & 0
     \end{bmatrix}}\quad
\textbf{G}_5 = \scaleG{\begin{bmatrix}
       0 & 0 & 0 & 0\\
       0 & 0 & 0 & 1\\
       0 & 0 & 0 & 0\\
       0 & 0 & 0 & 0
     \end{bmatrix}}\quad
\textbf{G}_6 =\scaleG{\begin{bmatrix}
       0 & 0 & 0 & 0\\
       0 & 0 & 0 & 0\\
       0 & 0 & 0 & 1\\
       0 & 0 & 0 & 0
     \end{bmatrix}}  
\end{align*}
The twist coordinates $\bm{\xi}$ are \textit{Lie-Cartan coordinates} of the first kind around the identity and relative to the basis $\{\textbf{G}_1,...,\textbf{G}_6\}$~\cite{varadarajan2013lie}.

Let $\bm{X}^c = [x_c,y_c,z_c]^T \in \mathbb{R}^3$ and $\bm{X}^w = [x_w,y_w,z_w]^T \in \mathbb{R}^3$ denote a 3D point  represented in camera and world frames respectively. Define
$\widebar{\bm{X}}^c = [x_c,y_c,z_c, 1]^T \in \mathbb{R}^4$ and $\widebar{\bm{X}}^w = [x_w,y_w,z_w, 1]^T \in \mathbb{R}^4$. With the left perturbation model, it is $\widebar{\bm{X}}^c = e^{\widehat{\bm{\xi}}} \textbf{T} \widebar{\bm{X}}^w$. By differentiating about the origin and using eq.~(\ref{eq:ExpDerivatives}), one has 
\begin{equation}\label{Eq:DerivativeX}
\frac{\partial \widebar{\bm{X}}^c}{\partial \xi_j} = \textbf{G}_j \widebar{\bm{X}}^c, \quad \frac{\partial \bm{X}^c}{\partial \xi_j} = \widetilde{\textbf{G}}_j \widebar{\bm{X}}^c, \quad \frac{\partial \bm{X}^c}{\partial (x_w,y_w,z_w)} = \textbf{R}
\end{equation} 
where $\widetilde{\textbf{G}}_j = [\textbf{I}_3,\bm{0}] \textbf{G}_j \in \mathbb{R}^{3\times 4}$   and $\textbf{R} \in SO(3)$ is the rotation matrix in $\textbf{T} \in SE(3)$. Overall, one obtains: 
\begin{align}
\label{Eq:FinalDerivativeX_A}
 \frac{\partial \bm{X}^c}{\partial \bm{\xi}} &=  \frac{\partial \bm{X}^c}{\partial \begin{bmatrix}
       \bm{\phi}^T & \bm{\rho}^T  
\end{bmatrix}^T} = 
 \begin{bmatrix} -\widehat{\bm{X}^c} & \textbf{I} \end{bmatrix}  \\
 \label{Eq:FinalDerivativeX_B}
 \frac{\partial \widebar{\bm{X}^c}}{\partial \bm{\xi}} &=  \frac{\partial \widebar{\bm{X}^c}}{\partial \begin{bmatrix}
       \bm{\phi}^T & \bm{\rho}^T  
\end{bmatrix}^T  } = 
  \begin{bmatrix} -\widehat{\bm{X}^c} & \textbf{I} \\
  \bm{0}^T & \bm{0}^T  
  \end{bmatrix} 
\end{align} 
where $\widehat{\bm{X}^c}\in \mathbb{R}^{3 \times 3}$ denotes the skew-symmetric matrix corresponding to $\bm{\bm{X}^c} \in \mathbb{R}^3$. 

Equations (\ref{Eq:FinalDerivativeX_A}--\ref{Eq:FinalDerivativeX_B}) can be alternatively obtained by considering a first order expansion of $\widebar{\bm{X}}^c$ after applying a left perturbation. In fact, by using eq.~(\ref{Eq:ExpFirstOrderApprox}), one has: 
\begin{multline}
\Delta \widebar{\bm{X}}^c = e^{\widehat{\bm{\xi}}} \textbf{T} \widebar{\bm{X}}^w - \textbf{T} \widebar{\bm{X}}^w \approx (\textbf{I} + \widehat{\bm{\xi}}) \textbf{T} \widebar{\bm{X}}^w 
- \textbf{T} \widebar{\bm{X}}^w \\
= \widehat{\bm{\xi}} \textbf{T} \widebar{\bm{X}}^w 
= \widehat{\bm{\xi}} \widebar{\bm{X}}^c =  \begin{bmatrix}
       \widehat{\bm{\phi}} & \bm{\rho} \\
       \bm{0}^T & 0  
\end{bmatrix} \widebar{\bm{X}}^c  \\
= \begin{bmatrix}
       \widehat{\bm{\phi}}\bm{X}^c  + \bm{\rho} \\
       0  
\end{bmatrix} = \begin{bmatrix}
       -\widehat{\bm{X}^c}\bm{\phi}  + \bm{\rho} \\
       0  
\end{bmatrix} = 
\begin{bmatrix}
       -\widehat{\bm{X}^c} & \textbf{I} \\
        \bm{0}^T & 0   
\end{bmatrix} \begin{bmatrix} \bm{\phi} \\ \bm{\rho}  \end{bmatrix}.
\end{multline}
This first order expansion (total differential) entails equations~(\ref{Eq:FinalDerivativeX_A}--\ref{Eq:FinalDerivativeX_B}).

\subsection{Camera Projection Jacobian}

The Jacobian of eq.~(\ref{eq:CameraProjection}) is:
\begin{equation}
\textbf{J}_c = \frac{\partial \bm{u}}{\partial\bm{X}^c} = \frac{\partial \bm{\gamma}(\textbf{K} \bm{X}^c)}{\partial\bm{X}^c} = \begin{bmatrix}
   \frac{f_x}{z_c} & 0 & -\frac{f_x x_c}{z_c^2} \\
   0 & \frac{f_y}{z_c} & -\frac{f_y y_c}{z_c^2}   
\end{bmatrix}
\end{equation}

\subsection{Jacobians of the Errors}

In this Section, we show how to compute the differential of the line reprojection/backprojection errors presented in Sect.~\ref{Sect:ReprojectionErrors}. 

We now focus on the 2D line reprojection error in eq.~(\ref{Eq:2DDistancePointLineSegment}) and  differentiate each of its two components by applying the calculus chain rule. 
One has  $d_{2D}(\cdot) = \bm{n}^T \bm{\gamma}(\textbf{K}\bm{X}^c) - h$ where $\bm{n}=[n_u,n_v]^T \in \mathbb{R}^2$ is constant. 
By differentiating $d_{2D}(\cdot)$ with respect to the components of $\bm{X}^c$, we obtain 
\begin{equation}
\frac{\partial d_{2D}}{\partial\bm{X}^c} = \textbf{J}_1 = \bm{n}^T \textbf{J}_c = 
\bigg[ \frac{f_x n_u}{z_c}, \frac{f_y n_v}{z_c}, - \frac{f_x n_u x_c}{z_c^2} - \frac{f_y n_v y_c}{z_c^2}\bigg].
\end{equation}
Using the previous equation with eq.~(\ref{Eq:DerivativeX}), we have 
\begin{equation}
\frac{\partial d_{2D}}{\partial \xi_j} = \frac{\partial d_{2D}}{\partial\bm{X}^c} \frac{\partial \bm{X}^c}{\partial \xi_j} = \textbf{J}_1 \widetilde{\textbf{G}}_j  \widebar{\bm{X}}^c
\end{equation}
and
\begin{equation}
\frac{\partial d_{2D}}{\partial(x_w,y_w,z_w)} = \frac{\partial d_{2D}}{\partial\bm{X}^c} \frac{\partial \bm{X}^c}{\partial (x_w,y_w,z_w)} =  \textbf{J}_1 \textbf{R} 
\end{equation}

We proceed in a similar way for eq.~(\ref{Eq:3DDistancePointBackProjPoint}). Define $\Delta\bm{X} \triangleq \bm{X}^c - \bm{\beta}_k(\bm{x})$ where $\bm{x} \in \Omega$ is the image point (on the $k$-th image) associated with $\bm{X}^c$. It is $d_P =\|\Delta \bm{X} \|$. Using eq.~(\ref{Eq:ModulusDerivative}) and eq.~(\ref{Eq:DerivativeX}), we have 
\begin{align}
\frac{\partial d_P}{\partial \xi_j} &= \frac{\partial d_P}{\partial \bm{X}^c} \frac{\partial \bm{X}^c}{\partial \xi_j} = \frac{\Delta\bm{X}^T }{\Vert \Delta\bm{X}\Vert }\widetilde{\textbf{G}}_j  \widebar{\bm{X}}^c \\ \frac{\partial d_P}{\partial(x_w,y_w,z_w)} &= \frac{\partial d_P}{\partial \bm{X}^c} \frac{\partial \bm{X}^c}{\partial (x_w,y_w,z_w)} = \frac{\Delta\bm{X}^T }{\Vert \Delta\bm{X}\Vert }\textbf{R}
\end{align}

We use the same approach for differentiating eq.~(\ref{Eq:3DDistancePointLineSegment}). Denote the backprojected points by $\bm{B}_P \triangleq \bm{\beta}_k(\bm{p}_i)$ and $\bm{B}_Q \triangleq \bm{\beta}_k(\bm{q}_i)$. These vectors are constant in the camera frame. 
Let $\Delta \bm{B} \triangleq  \bm{B}_P -  \bm{B}_Q$ be the constant vector difference between them. Define $\Delta\bm{P} \triangleq \bm{X}^c - \bm{B}_P $, $\Delta\bm{Q} \triangleq \bm{X}^c - \bm{B}_Q$ and $\bm{V} \triangleq (\bm{X}^c - \bm{B}_P)\times (\bm{X}^c - \bm{B}_Q)$. It is $d_{3D} = \| \bm{V} \|/\|\Delta \bm{B} \|$. Using eq.~(\ref{Eq:CrossProductDerivative}) and considering that  $\Delta \bm{B}$ is a constant vector, one has from eq.~(\ref{Eq:3DDistancePointLineSegment})
\begin{equation}
\frac{\partial d_{3D}}{\partial \bm{X}^c} = \frac{\bm{V}^T ( \widehat{\Delta\bm{P}}-\widehat{\Delta\bm{Q}})}{ \Vert \bm{V} \Vert \|\Delta \bm{B} \|} = -\frac{\bm{V}^T \widehat{\Delta \bm{B}}}{\Vert \bm{V} \Vert \|\Delta \bm{B} \|}.
\end{equation}
Consequently, considering eq.~(\ref{Eq:DerivativeX}), it follows
\begin{align}
\Scale[1]{
\frac{\partial d_{3D}}{\partial \xi_j} }&= 
\Scale[1]{\frac{\partial d_{3D}}{\partial \bm{X}^c} \frac{\partial \bm{X}^c}{\partial \xi_j} } = 
\Scale[1]{-\frac{\bm{V}^T \widehat{\Delta \bm{B}}}{\Vert \bm{V} \Vert \|\Delta \bm{B} \|}\widetilde{\textbf{G}}_j \widebar{\bm{X}}^c} \\
\Scale[1]{
\frac{\partial d_{3D}}{\partial(x_w,y_w,z_w)} } &= 
\Scale[1]{\frac{\partial d_{3D}}{\partial \bm{X}^c} \frac{\partial \bm{X}^c}{\partial (x_w,y_w,z_w)} } = \Scale[1]{-\frac{\bm{V}^T \widehat{\Delta \bm{B}}}{\Vert \bm{V} \Vert \|\Delta \bm{B} \| }\textbf{R}}.
\end{align}
Now, define the distance $d_B(i, k,\bm{X}^w) \triangleq d_{3D}(i, k,\bm{X}^w) + \mu d_P(\bm{x}, k,\bm{X}^w)$, which is one of the component of the distance vector $\bm{d}_B$ in eq.~(\ref{Eq:3DDistanceBB}). One has
\begin{align}
\Scale[1]{ 
\frac{\partial d_B(i, k,\bm{P}^w)}{\partial \xi_j} } &= 
\Scale[1]{ \big( -\frac{\bm{V}_P^T \widehat{\Delta \bm{B}}}{\Vert \bm{V}_P \Vert \|\Delta \bm{B} \| } +  \mu \frac{(\bm{P}^c - \bm{B}_P)^T }{\Vert \bm{P}^c - \bm{B}_P \Vert }\big) \widetilde{\textbf{G}}_j  \widebar{\bm{P}}^c } \\
\Scale[1]{ 
\frac{\partial d_B(i, k,\bm{Q}^w)}{\partial \xi_j} } &= 
\Scale[1]{ \big( -\frac{\bm{V}_Q^T \widehat{\Delta \bm{B}}}{ \Vert \bm{V}_Q \Vert \|\Delta \bm{B}\| } +  \mu \frac{(\bm{Q}^c - \bm{B}_Q)^T }{\Vert \bm{Q}^c - \bm{B}_Q \Vert }\big)\widetilde{\textbf{G}}_j  \widebar{\bm{Q}}^c }
\end{align}
where $\bm{V}_P$ and $\bm{V}_Q$ are obtained from $\bm{V}$ by replacing $\bm{X}^c$ respectively with $\bm{P}^c$ and $\bm{Q}^c$. In the same way, we obtain 
\begin{align}
\Scale[1]{ 
\frac{\partial d_B(i, k,\bm{P}^w)}{\partial(x^p_w,y^p_w,z^p_w)} } &= 
\Scale[1]{ \big( -\frac{\bm{V}_P^T \widehat{\Delta \bm{B}}}{\Vert \bm{V}_P \Vert \|\Delta \bm{B} \|} +  \mu \frac{(\bm{P}^c - \bm{B}_P)^T }{\Vert \bm{P}^c - \bm{B}_P \Vert }\big) \textbf{R}  } \\
\Scale[1]{ 
\frac{\partial d_B(i, k,\bm{Q}^w)}{\partial(x^q_w,y^q_w,z^q_w)} } &= 
\Scale[1]{ \big( -\frac{\bm{V}_Q^T \widehat{\Delta \bm{B}}}{\Vert \bm{V}_Q \Vert \|\Delta \bm{B} \| } +  \mu \frac{(\bm{Q}^c - \bm{B}_P)^T }{\Vert \bm{Q}^c - \bm{B}_Q \Vert }\big) \textbf{R}  } 
\end{align}
where $\bm{P}^w \triangleq [x^p_w,y^p_w,z^p_w]^T$ and $\bm{Q}^w \triangleq [x^q_w,y^q_w,z^q_w]^T$.

\subsection{Covariances}

In this section, we sketch how to compute the covariances presented in Sect.~\ref{Sect:ReprojectionErrors}.

We first focus on the covariance $\bm{\Sigma}_{li}$ of eq.~(\ref{Eq:SigmaD2DLine}). With reference to eq.~(\ref{Eq:LineEquation}), let $ \bm{l} \triangleq [q_v - p_v, p_u - q_u ] \in \mathbb{R}^2$ be a vector normal to the line segment joining the endpoints $(\bm{p},\bm{q})$ and let $\bm{n} = \frac{\bm{l} }{\| \bm{l}\|} \in \mathbb{R}^2$ be the corresponding unit normal. Consider the scalar 2D distance used in eq.~(\ref{Eq:2DDistancePointLineSegment}), i.e. $d_{2D} = \bm{n}^T [u,v]^T - h$, where $h = \bm{n}^T[p_u,p_v]^T$ and $\bm{u}=[u,v]= \bm{\gamma}(\textbf{K}[\textbf{R}_k,\bm{t}_k] \widebar{\bm{X}}^w)$ is the camera projection of the map point $\bm{X}^w$.  We compute the covariance $\sigma_{d2D}$ in eq.~(\ref{Eq:SigmaD2DLine}) by using the calculus chain rule, i.e. by first considering $d_{2D}$ as a function of $\bm{n}$ and $h$ instead of $(\bm{p},\bm{q})$. One has
\begin{equation}
\sigma_{d2D}^2 = \frac{\partial d_{2D}}{\partial (n_u,n_v,h)}  
\begin{bmatrix}
       \bm{\Sigma}_n & \bm{0}           \\
       \bm{0} & \sigma_h^2
     \end{bmatrix}
\frac{\partial d_{2D}}{\partial (n_u,n_v,h)}^T     
\in \mathbb{R}
\end{equation} 
where $\bm{\Sigma}_n  \in \mathbb{R}^{2 \times 2}$ is the covariance matrix of the normal $\bm{n}$, $\sigma_h^2 \in \mathbb{R}$ is the variance of the signed distance $h$ and $\frac{\partial d_{2D}}{\partial (n_u,n_v,h)} = [u, v, -1]$. Hence, it is
\begin{equation}\label{Eq:SigmaD2d}
\sigma_{d2D}^2 = \begin{bmatrix}
       u  & v
     \end{bmatrix} \bm{\Sigma}_n \begin{bmatrix}
       u        \\
       v
     \end{bmatrix}  + \sigma_h^2
\end{equation} 

We now show hot to compute   
$ \bm{\Sigma}_n$ and $\sigma_h^2$. Given that $\bm{n} = \bm{n}(p_u,p_v, q_u,q_v)$ and $h=h(p_u,p_v, q_u,q_v)$, one has
\begin{align}
\bm{\Sigma}_n &= \frac{\partial \bm{n}}{\partial (\bm{p},\bm{q})} \sigma_{li}^2 \textbf{I}_4  \frac{\partial \bm{n}}{\partial (\bm{p},\bm{q})}^T \\
\label{Eq:SigmaH}
\sigma_h^2 &= \frac{\partial h}{\partial (\bm{p},\bm{q})} \sigma_{li}^2 \textbf{I}_4  \frac{\partial h}{\partial (\bm{p},\bm{q})}^T.
\end{align} 
where $\frac{\partial \bm{n}}{\partial (\bm{p},\bm{q})} \in \mathbb{R}^{2\times 4}$ and $\frac{\partial h}{\partial (\bm{p},\bm{q})} \in \mathbb{R}^{1\times 4}$ are Jacobians with respect to both image endpoints coordinates.

Using eq.~(\ref{Eq:NormalDerivative}), one has 
\begin{equation}\label{Eq:SigmaN}
\bm{\Sigma}_n = \frac{1}{\|\bm{l}\|^2} \Big( \textbf{I}_2 - \bm{n} \bm{n}^T \Big)\frac{\partial \bm{l}}{\partial (\bm{p},\bm{q})} \sigma_{li}^2 \textbf{I}_4  \frac{\partial \bm{l}}{\partial (\bm{p},\bm{q})}^T \Big( \textbf{I}_2 - \bm{n} \bm{n}^T \Big)
\end{equation}
where 
\begin{equation}
\frac{\partial \bm{l}}{\partial (\bm{p},\bm{q})} = \begin{bmatrix}
0 & -1 & 0  & 1 \\
1 &  0 & -1 & 0 
\end{bmatrix}.
\end{equation}
Then, using again eq.~(\ref{Eq:NormalDerivative}), it follows
\begin{align}\label{Eq:Dhdpq}
\frac{\partial h}{\partial (\bm{p},\bm{q})} &= \frac{\bm{l}^T }{\| \bm{l}\|} \frac{\partial }{\partial (\bm{p},\bm{q})} 
\begin{bmatrix} p_u \\ p_v \end{bmatrix} \nonumber \\ 
&\quad +\Bigg( \frac{1}{\| \bm{l}\|} \frac{\partial \bm{l}}{\partial (\bm{p},\bm{q})}^T\Big( \textbf{I}_2 - \bm{n} \bm{n}^T \Big) \begin{bmatrix} p_u \\ p_v \end{bmatrix} \Bigg)^T
\end{align}
where 
\begin{equation}
\frac{\partial }{\partial (\bm{p},\bm{q})}\begin{bmatrix} p_u \\ p_v \end{bmatrix}  = \begin{bmatrix}
1 & 0 & 0 & 0 \\
0 & 1 & 0 & 0 
\end{bmatrix}.
\end{equation}
Plugging eq.~(\ref{Eq:Dhdpq}) into eq.~(\ref{Eq:SigmaH}), one obtains $\sigma_h^2$.
Finally, plugging eq.~(\ref{Eq:SigmaN}) and (\ref{Eq:SigmaH}) into eq.~(\ref{Eq:SigmaD2d}), one obtains $\sigma_{d2D}$.

We now focus on the covariance $\bm{\Sigma}_{Li}$ in eq.~(\ref{Eq:SigmaLi}). In order to compute it, we need to differentiate the 3D line-line distance $d_{3D}$ and the 3D point-point distance $d_P$ with respect to both $\bm{B}_P = \bm{\beta}_k(\bm{p}_i)$ and $\bm{B}_Q = \bm{\beta}_k(\bm{q}_i)$. We adopt the same notation used in previous subsection. Recall that $d_{3D} = \| \bm{V} \|/\|\Delta \bm{B} \|$ and $d_P = \| \Delta \bm{X} \|$. Using eq.~(\ref{Eq:ModulusDerivative}--\ref{Eq:InvModulusDerivative}), one has
\begin{align}\label{Eq:Dd3DdBp}
\frac{\partial d_{3D}}{\partial \bm{B}_P} &= \frac{\bm{V}^T}{\Vert \bm{V} \Vert \|\Delta \bm{B} \| }\frac{\partial \bm{V}}{\partial \bm{B}_P} + \Vert \bm{V} \Vert \frac{\partial}{\partial \bm{B}_P} \frac{1}{\|\Delta \bm{B} \|} = \nonumber \\
\quad &= \frac{\bm{V}^T}{\Vert \bm{V} \Vert \|\Delta \bm{B} \| }\widehat{\Delta \bm{Q}} - \Vert \bm{V} \Vert \frac{\Delta \bm{B} ^T}{\|\Delta \bm{B} \|^3}
\end{align}
\begin{align}\label{Eq:Dd3DdBq}
\frac{\partial d_{3D}}{\partial \bm{B}_Q} &= \frac{\bm{V}^T}{\Vert \bm{V} \Vert \|\Delta \bm{B} \| }\frac{\partial \bm{V}}{\partial \bm{B}_Q} + \Vert \bm{V} \Vert \frac{\partial}{\partial \bm{B}_Q} \frac{1}{\|\Delta \bm{B} \|} = \nonumber \\
\quad &= -\frac{\bm{V}^T}{\Vert \bm{V} \Vert \|\Delta \bm{B} \| }\widehat{\Delta \bm{P}} + \Vert \bm{V} \Vert \frac{\Delta \bm{B} ^T}{\|\Delta \bm{B} \|^3}
\end{align}
and
\begin{equation}\label{Eq:DPdBx}
\frac{\partial d_P}{\partial \bm{B}_h} = -\frac{\Delta\bm{X}_h^T }{\Vert \Delta\bm{X}_h\Vert }
\end{equation}
where $\Delta\bm{X}_h \triangleq \bm{X}^c - \bm{B}_h$ and $h \in \{P,Q\}$.
Starting from eq.~(\ref{Eq:SigmaBeta1}), it is  easy to show that 
\begin{equation}\label{Eq:SigmaBetaComputed}
\bm{\Sigma}_{\beta}(\bm{x}) =
\begin{bmatrix}
       \sigma_{z}^2 x^2+ \frac{\delta^2 \sigma_{li}^2}{f_x^2} & \sigma_{z}^2 x y & \sigma_{z}^2 x \\
       \sigma_{z}^2 x y & \sigma_{z}^2 y^2+ \frac{\delta^2 \sigma_{li}^2}{f_y^2} & \sigma_{z}^2 y \\       
       \sigma_{z}^2 x & \sigma_{z}^2 y & \sigma_{z}^2  
\end{bmatrix}
\end{equation}
where $\bm{x} = [u,v]^T \in \Omega$ is a line endpoint on the image plane and $\bm{\beta}_k(\bm{x}) = [x,y,\delta] \in \mathbb{R}^3$ is its backprojection. Combining eq.~(\ref{Eq:SigmaD3D}) with eq.~(\ref{Eq:Jd3D}), one obtains 
\begin{align}\label{Eq:SigmaD3DNewA}
\Scale[0.9]{
\sigma_{dB}^2(i, k,\bm{P}^w)} &= \Scale[0.9]{(\frac{\partial  d_{3D}}{\partial \bm{B}_P}+\mu \frac{\partial  d_P}{\partial \bm{B}_P})\Big|_{\bm{P}^w}
 \bm{\Sigma}_{\beta}(\bm{p}_i)
(\frac{\partial d_{3D}}{\partial \bm{B}_P}+\mu \frac{\partial  d_P}{\partial \bm{B}_P})^T\Big|_{\bm{P}^w} + } \nonumber \\ 
\quad &+ \Scale[0.9]{
 \frac{\partial  d_{3D}}{\partial \bm{B}_Q}\Big|_{\bm{P}^w}          
 \bm{\Sigma}_{\beta}(\bm{q}_i)
\frac{\partial d_{3D}}{\partial \bm{B}_Q}^T\Big|_{\bm{P}^w}} 
 \\ \label{Eq:SigmaD3DNewB}
\Scale[0.9]{
\sigma_{dB}^2(i, k,\bm{Q}^w)} &= \Scale[0.9]{\frac{\partial  d_{3D}}{\partial \bm{B}_P}\Big|_{\bm{Q}^w}          
 \bm{\Sigma}_{\beta}(\bm{p}_i)
\frac{\partial d_{3D}}{\partial \bm{B}_P}^T\Big|_{\bm{Q}^w} +} \nonumber \\ 
\quad &+ \Scale[0.9]{
 (\frac{\partial  d_{3D}}{\partial \bm{B}_Q}+\mu \frac{\partial  d_P}{\partial \bm{B}_Q})\Big|_{\bm{Q}^w}
 \bm{\Sigma}_{\beta}(\bm{q}_i)
(\frac{\partial d_{3D}}{\partial \bm{B}_Q}+\mu \frac{\partial  d_P}{\partial \bm{B}_Q})^T\Big|_{\bm{Q}^w}
}
\end{align}
where we used the facts that $\frac{\partial  d_P}{\partial \bm{B}_Q}\Big|_{\bm{P}^w}=0$ and $\frac{\partial  d_P}{\partial \bm{B}_P}\Big|_{\bm{Q}^w}=0$.
Plugging eqs.~(\ref{Eq:Dd3DdBp}), (\ref{Eq:Dd3DdBq}) and~(\ref{Eq:SigmaBetaComputed}) into eq.~(\ref{Eq:SigmaD3DNewA}-\ref{Eq:SigmaD3DNewB}), we obtain $\bm{\Sigma}_{Li}$ in eq.~(\ref{Eq:SigmaLi}).
